\documentclass[lettersize,journal]{IEEEtran}
\usepackage{amsmath,amsfonts}
\usepackage{float} 
\usepackage[ruled]{algorithm2e}
\usepackage{array}
\usepackage[caption=false,font=scriptsize,labelfont=sf,textfont=sf]{subfig}
\usepackage{textcomp}
\usepackage{stfloats}
\usepackage{url}
\usepackage{verbatim}
\usepackage{graphicx}
\usepackage{cite}
\usepackage{xcolor}
\usepackage{setspace}
\usepackage{array}
\usepackage{bm}
\usepackage{multirow}
\usepackage{soul}
\newcommand\zy[1]{\textcolor{black}{{#1}}}
\newcommand\oy[1]{\textcolor{black}{{#1}}}
\newcommand\mr[1]{\textcolor{black}{{#1}}}

\begin{document}
\newtheorem{definition}{Definition}[section]
\newtheorem{proposition}{Proposition}[section]
\title{Debiasing Graph Representation Learning based on Information Bottleneck}

\author{Ziyi Zhang$^\dagger$, Mingxuan Ouyang$^\dagger$,
Wanyu Lin$^*$,
Hao Lan,
Lei Yang
\thanks{\textit{Mingxuan Ouyang and Ziyi Zhang contributed equally to this work, denoted with $\dagger$; the corresponding author is Wanyu Lin denoted with *. This work was done when Ziyi Zhang was an intern at The Hong Kong Polytechnic University.}}
\thanks{Ziyi Zhang is with the School of Software Engineering, South China University of Technology, Guangzhou, China, and The Hong Kong Polytechnic University, Hong Kong, China (email: seziyizhang@mail.scut.edu.cn).}
\thanks{Wanyu Lin and Mingxuan Ouyang are with the Department of Computing, The Hong Kong Polytechnic University, Hong Kong, China (e-mail: wan-yu.lin@polyu.edu.hk, mingxuan23.ouyang@connect.polyu.hk).}
\thanks{Hao Lan is with the Department of Computer Science and Technology, Tsinghua University, Beijing, China (e-mail: lanhao@mail.tsinghua.edu.cn).}
\thanks{Lei Yang is with the School of Software Engineering, South China University of Technology, Guangzhou, China (email: sely@scut.edu.cn).}}


\IEEEpubid{0000--0000/00\$00.00~\copyright~2024 IEEE}

\maketitle

\begin{abstract}
Graph representation learning has shown superior performance in numerous real-world applications, such as finance and social networks. Nevertheless, most existing works might make discriminatory predictions due to insufficient attention to fairness in their decision-making processes. This oversight has prompted a growing focus on fair representation learning. Among recent explorations on fair representation learning, prior works based on adversarial learning usually induce unstable or counterproductive performance. To achieve fairness in a stable manner, we present the design and implementation of GRAFair, a new framework based on a variational graph auto-encoder. The crux of GRAFair is the Conditional Fairness Bottleneck, where the objective is to capture the trade-off between the utility of representations and sensitive information of interest. By applying variational approximation, we can make the optimization objective tractable. Particularly, GRAFair can be trained to produce informative representations of tasks while containing little sensitive information without adversarial training. Experiments on various real-world datasets demonstrate the effectiveness of our proposed method in terms of fairness, utility, robustness, and stability. 

\end{abstract}




\section{Introduction}
\IEEEPARstart{G}{raph} neural networks (GNNs) have demonstrated increasing popularity for learning over graph-structured data, such as social networks \cite{ying2018graph, lin2021medley}, knowledge graphs \cite{gao2020deep}, and chemical molecules \cite{gilmer2017neural}. They have exhibited impressive performance in aggregating information and learning representations following a message-passing paradigm from both the node features and the graph structure. The learned representations can be used in various graph tasks such as graph classification, node classification, and link prediction \cite{ wang2024multi, lin2020adversarial, lin2020shoestring}.
As numerous variants of GNNs are increasingly being implemented in various ethics-critical domains, including recommendation systems \cite{ying2018graph, lin2024graph}, financial prediction \cite{yang2020financial}, and diagnostic medical \cite{ahmedt2021graph}. It is essential to ensure the integrity and reliability of the predictive models. However, recent research has shown that these high-performance GNNs usually overlook fairness issues \cite{dai2021say, dong2022edits,mehrabi2021survey, wang2024generating}, \oy{which limits model adoption in many real-world applications.}

\oy{The biases in GNN predictions are typically attributed to a combination of node attributes and graph structure. Firstly, GNNs can capture the statistical correlation between nodes' raw attributes and sensitive attributes, leading to the leakage of sensitive information in encoded representations \cite{ling2022learning}. Secondly, message-passing in GNNs can exacerbate sensitive biases in node representations within the same sensitive group due to homophily effects \cite{mcpherson2001birds, yang2024fairsin, wu2020comprehensive}. Homophily effects refer to the tendency of nodes with similar sensitive attributes to link with each other \cite{la2010randomization}. As a result, biased node representation might lead to severe discrimination in the downstream task.}


\IEEEpubidadjcol
\begin{figure}[!t]
\centering
\includegraphics[width=3in]{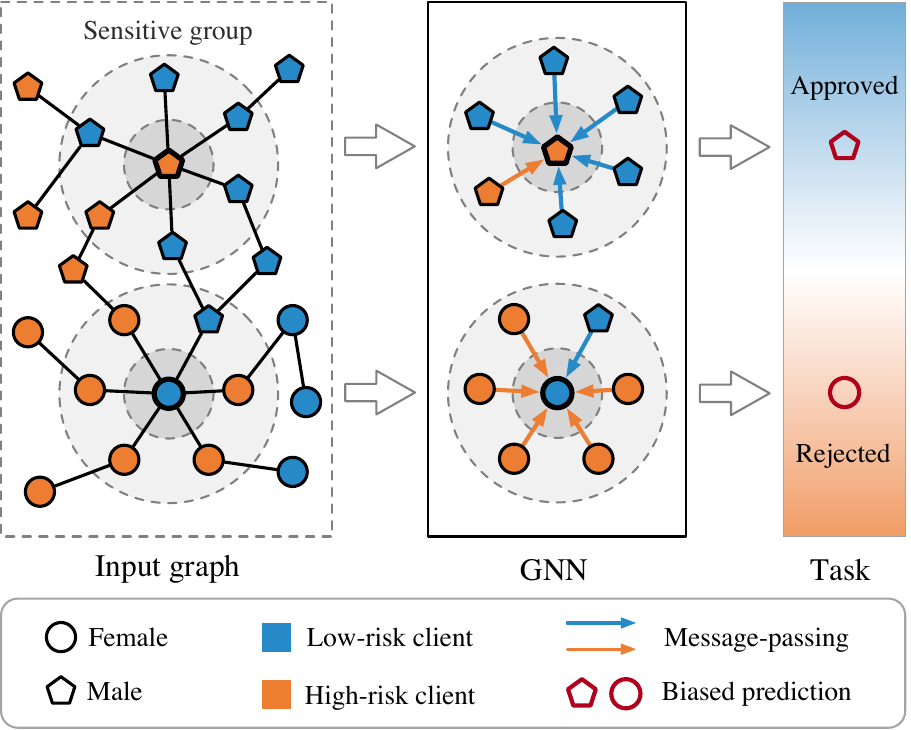}
\caption{An illustration of gender bias in GNN for loan approval prediction.}

\label{fig_1}
\end{figure}

Take loan approval prediction as an example: clients are treated as nodes, forming a graph based on clients' similar payment behaviors \cite{yeh2009comparisons, ma2022learning}. The goal is to predict whether a client's loan request will be approved. Naturally, this task can be formulated as a node classification problem and solved with GNNs. \oy{As shown in Figure \ref{fig_1}, clients within the same sensitive group tend to have similar representations due to the message-passing mechanism in the GNN. The homophily effects lead to GNN inadvertently perpetuating or amplifying gender bias against these sensitive groups. Therefore, as illustrated in Figure \ref{fig_1}, a female with actually low repayment risk is rejected, while a male with actually high repayment risk is approved. The discriminatory predictions favoring males indicate that gender in loan approval tasks is over-weighted. In banking decision-making processes, it is essential to prioritize attributes that substantially impact loan repayment capacity, such as occupation and revenue.}
\oy{A straightforward debiasing approach involves removing sensitive attributes from all nodes, but this might not effectively improve fairness. This is because biases can still arise from the correlation between sensitive attributes and other attributes or graph structure \cite{mehrabi2021survey}. }

To tackle the above problems, fair representation learning has been proposed and widely studied due to its generality. \oy{The high-level goal of fair representation learning is to learn an encoding function that maps nodes in a graph to unbiased representations, retaining task-related information while being independent of sensitive attributes.}
\oy{Many recent fair representation learning approaches have introduced fairness considerations into GNNs\cite{wu2021learning,agarwal2021towards,ma2022learning,dai2021say,wang2022improving}. Most of the state-of-the-art methods are applying heuristic\cite{agarwal2021towards} or based on adversarial learning by playing a max-min game between a fair encoder and an adversary of sensitive attributes \cite{ma2022learning,dai2021say,wang2022improving}.} 
Other approaches weigh up performance against fairness, such as revising features \cite{zhang2017achieving,d2017conscientious}, fairness regularization by adding a penalty term \cite{buyl2021kl,liu2022fair}, and disentanglement \cite{creager2019flexibly,oh2022learning}. However, these approaches may result in unstable or counter-productive performance via adversarial training \cite{arjovsky2017towards, oh2022learning}, or cannot be directly applied in GNNs for the absence of simultaneous consideration of bias from node attributes and graph structures. 
While these methods have shown compelling results in enhancing fairness, a high cost due to instability or drastically reduced utility limits their widespread use in practice. From the perspective of real-life applications, it is essential to capture the trade-off between fairness and utility while maintaining the stability of models.

To this end, we propose a new framework based on the variational auto-encoding architecture \cite{kipf2016variational}, GRAFair (\textbf{G}raph \textbf{R}epresentation Le\textbf{A}rning based on \textbf{Fair}ness Information Bottleneck), to address the aforementioned problem. Our framework is optimized with Conditional Fairness Bottleneck (CFB) \cite{rodriguez2021variational} tackled with a variational approach. Specifically, CFB aims to encode the graph data into fair representations by maintaining the information relevant to the task while minimizing the sensitive and irrelevant information. \oy{It gives us a principle to evaluate the utility and fairness of the model from an information-theoretic perspective. By this principle, we derive the upper and lower bounds for solving this optimization problem to achieve optimal trade-off between utility and fairness.}
However, there are two major challenges to solving the problem on the graphs: i) Graph fair representation learning is still a developing research area due to the challenges introduced by complexity and non-Euclidean graph structure; ii) The common approaches to solving the optimization problem is intractable, making it difficult to obtain an optimal solution. \oy{To address the above challenges, we apply variational approximation and use the reparameterization trick to convert the optimization problem into a tractable one.} Our major contributions are as follows:


\begin{itemize}
\item We propose GRAFair, a new framework designed to learn fair representations based on a variational graph auto-encoder. It aims to alleviate the impact of sensitive and irrelevant information on downstream tasks \oy{in a stable manner.}

\item We address the key challenge of intractable optimization on graphs considering the Information Bottleneck in fair representation learning, and our theoretical analyses show that such a problem can be efficiently solved.

\item We perform extensive experiments on three real-world datasets, and the results show that GRAFair outperforms the state-of-the-art baselines on fairness, utility, robustness and stability.

\end{itemize}

\section{Related work}
\label{relatedwork}

\subsection{Graph neural networks}Graph Neural Networks (GNNs) convert a graph into low-dimensional node representations informative of attributes and structure used for graph downstream tasks. Various architectures are proposed to solve graph representation learning, such as Graph Convolutional Network (GCN)\cite{kipf2016semi}, GraphSAGE\cite{hamilton2017inductive} and Graph Attention Network (GAT)\cite{velivckovic2017graph}. These methods generate representations by combining the features from neighbors following an information aggregation scheme \zy{with different neighborhood sampling and aggregation approaches.}

In this work, we propose a novel framework founded on a variational graph auto-encoder (VGAE) \cite{kipf2016variational} for fair representation learning. The model is based on the concept of variational autoencoders (VAEs), which are generative models that incorporate probabilistic encoding. Diverging from traditional GNNs, VGAE introduces a probabilistic dimension to estimate a posterior distribution by inferring latent variables given the observations, and it enriches node representations with informativeness by capturing both node features and graph structure.

\subsection{Fair representation learning on graphs} 
Fair representation learning on graphs aims to learn a representation that can be used for downstream tasks without the impact of sensitive information while maintaining utility. There are two primary considerations for fairness. Group fairness requires predictions among each group of nodes made by GNNs to be independent with sensitive attributes (e.g., gender, race), including demographic parity \cite{wan2021modeling} and equal opportunity \cite{dong2022fairness}. Individual fairness emphasizes that individuals should have similar predictions if they have similar attributes evaluated by similarity metrics (e.g., distance metrics), whether they belong to the same sensitive attribute group or not \cite{wan2021modeling}, even in the case of counterfactual \cite{ma2022learning}. 


\oy{Additionally, fair representation learning on graphs needs to consider the homophily effects \cite{la2010randomization}. Message aggregation among clients/nodes with the same sensitive attribute further segregates the representations between different sensitive groups, which magnifies the association of their predictions with sensitive attributes \cite{zhu2023fairness, agarwal2021towards}.
As a result, the prediction made by GNNs might be biased against sensitive information (\textit{e}.\textit{g}., gender), which indicates that GNNs are vulnerable to unfairness due to sensitive attributes of interest in data. In other words, without limiting such sensitive attributes, the model could make discriminatory predictions due to discrimination and societal bias in historical data and magnification of GNN message passing \cite{dai2021say, wu2020comprehensive, yang2024fairsin}.}


Several methods have proposed different strategies to achieve fair graph representation learning. Some adversarial learning-based approaches involve training an adversary model to predict sensitive attributes from node embeddings and subsequently updating the node embeddings to minimize the prediction accuracy of the adversary \cite{dai2021say, wang2022improving}.
Some approaches directly add fairness constraints to the objective function of the machine learning models, which work as regularization terms to balance the performance in prediction and fairness \cite{ li2021dyadic, kang2020inform}. 
\oy{In NIFTY \cite{agarwal2021towards}, employing the counterfactual regularization to perturb node attributes and drop edges may result in the loss of non-sensitive information.} 
There are also methods that concentrate on debiasing the original graph data, including the elimination of attribute bias and structural bias. These techniques involve adjustments to the feature matrix and adjacency matrix, ensuring a fair distribution of attributes and unbiased propagation of information \cite{dong2022edits, spinelli2021fairdrop}. \oy{Furthermore, some fairness works are based on other architectures rather than GNNs. For example, FairGT \cite{luo2024fairgt} is tailored for graph transformers, integrating an eigenvector selection mechanism to enhance the graph information encoding and a fairness-aware node feature encoder to keep the independence of sensitive attributes.}

\subsection{Adversarial learning} With a significant improvement in generative adversarial learning (GAN) \cite{arjovsky2017towards}, adversarial training methods have been widely utilized for fair representation learning \cite{dai2021say, wang2022improving}. In general, the adversarial fair representation learning framework consists of two parts, a fair encoder and an adversary. The fair encoder tries to generate representations that filter out sensitive information, while the adversary attempts to identify sensitive attributes of generated representations. The objective goal can be considered as a minimax optimization. During the training process, the encoder and the adversary both use information from each other iteratively to improve their own model.

\oy{There are several recent methods that employ adversarial training to filter out sensitive biases.} 
\oy{Wang {\em et al.}~ \cite{wang2022improving} proposed FairVGNN to learn fair node representations by masking sensitive-correlated channels and clamping weights, aided by adversarial discriminators. Ma {\em et al.}~ \cite{ma2022learning} designed GEAR based on adversarial learning to meet fairness requirements by minimizing the discrepancy between the representations learned from the original graph and those from counterfactual augmentation graphs. Dai {\em et al.}~ \cite{dai2021say} deployed an adversary that can infer sensitive attributes to ensure GNNs make predictions independent of the estimated sensitive information.
Graphair \cite{ling2022learning} is an automated graph augmentation method based on adversarial learning, employing an adversary model to predict sensitive attributes from the augmented graph. It aims to preserve useful information from original graphs via contrastive learning, maximizing the agreement between original and augmented graphs.}

Although adversarial learning has achieved good performance, there exist two main problems: one is asynchronous convergence between two parts, and the other is the vanishing gradient. 
\oy{The above approaches based on adversarial learning suffer from convergence instability and vanishing gradient problems, making their performance unstable and even counter-productive \cite{arjovsky2017towards, oh2022learning}. Therefore, we propose GRAFair without adversarial learning to enhance fairness while maintaining the stability and utility of GNNs in downstream tasks.}

\section{Preliminaries}
\label{preliminaries}
\subsection{Notations}


\mr{We denote an attributed graph by $\mathcal{G}=(\mathcal{V},\mathcal{E})$ with $n$ nodes, where $\mathcal{V}=\{v_1,v_2,..,v_n\}$ is the node set, $\mathcal{E}=\{(i,j):i,j \in \mathcal{V}\}$ is the edge set, and each $(i,j)$ represents an edge connecting node $i$ and $j$.
 Let $\mathbf{A}\in\mathbb{R}^{n\times{n}}$ denote the adjacency matrix of $\mathcal{G}$, where $\mathbf{A}_{ij}=1$ if $(i,j)\in\mathcal{E}$ or $0$ otherwise. $\mathcal{G}$ is assumed to be undirected and unweighted, but it is naturally extended to be directed or weighted.
$\mathbf{X}=[\mathbf{x}_1,\cdots,\mathbf{x}_n]^\top\in\mathbb{R}^{n\times d}$ is the node feature matrix, where $\mathbf{x}_{i} \in \mathbb{R}^{1\times d}$ represents the feature vector for node $v_i$. We use $\mathbf{S}\in \{0,1\}^n$ to denote binary sensitive attributes of all nodes, where the $i$-th element of $\mathbf S$, \textit{i}.\textit{e.}, $s_i\in \{0,1\}$ is included in $\mathbf{x}_i$. The learned representation of the node $v_i$ is denoted as $\mathbf{z}_i\in \mathbf{Z}$. In this paper, We focus on binary node classification task, \textit{i}.\textit{e.}, the node $v_i$ has the label $y_i\in\{0,1\}$, and $\mathbf Y \in \{0,1\}^n$ is the label vector.}

\subsection{Graph message passing}
Graph Neural Networks (GNNs) have gained increasing attention due to their capacity to use both node attributes and graph structure for representation learning. GNNs are predicated based on a graph message-passing mechanism, wherein both node attributes and graph structure contribute to learning a representation $\mathbf{z}_i$ for each node $v_i\in\mathcal{V}$ \cite{zhou2020graph}.
Different GNNs follow different approaches to neighborhood aggregation. For a GNN with $L$ layers, the final node representations in the last layer can capture the structural information and aggregate information from its $L$-hop neighbors \cite{wu2020comprehensive}. Typically, iterative message-passing is a dynamic mechanism where nodes learn their representations by aggregating information over successive layers and enable GNNs to capture dependencies in a graph, which can be formulated as follows:
\begin{equation}
\begin{aligned}
\label{mesgpassing}
&\mathbf{a}_{i}^{k} = \mathbf{Aggregate}(\{\mathbf{h}_{i}^{k-1}:i\in \mathcal{N}(i)\}),\\& 
\mathbf{h}_{i}^{k} = \mathbf{Combine}(\mathbf{h}_{i}^{k-1}, \mathbf{a}_{i}^{k}),
\end{aligned}
\end{equation} 

where $\mathbf{a}_{i}^{k}$ is aggregated information from neighbors after $k$ iterations, $\mathbf{h}_{i}^{k}$ is the embedding of the node $v_i\in\mathcal{V}$ at the $k$-th layer, \oy{$ \mathcal{N}(i) $ is the set of neighbors of $i$}, and the representation 
$\mathbf{z}_{i} = \mathbf{h}_{i}^{L}$.

 \subsection{Information bottleneck}

 \mr{Information Bottleneck (IB) \cite{tishby99information} provides a critical information-theoretic principle for pattern analysis and representation learning, and has gained widespread popularity. 
 It formulates an information compression model to learn optimal representations that contain the minimal sufficient information for the downstream task. 
 On one hand, IB distinguishes different input samples and encourages the representation to be maximally informative for high accuracy. On the other hand, IB compresses similar inputs and discourages the representation from acquiring irrelevant information for downstream tasks \cite{hu2024survey, hu2023multiview}.
 Therefore, the learned model based on IB naturally avoids overfitting on numerous practical tasks and becomes more robust to adversarial attacks \cite{wu2020graph}.
 Specifically, IB maximally compresses the input data $\mathbf X$ into a compact representation $\mathbf Z$ while preserving sufficient relevant information about prediction task $\mathbf Y$. This can be formulated by: 
 \begin{equation}
\min\limits_{\rm P(\mathbf{Z}|\mathbf{X})}I(\mathbf Z;\mathbf X) -\lambda I(\mathbf Z;\mathbf Y),
\end{equation}
where $\lambda$ is the trade-off between irrelevant information and preserved information.
 }

 
 \subsection{Conditional fairness bottleneck}

 \mr{Conditional Fairness Bottleneck (CFB) \cite{rodriguez2021variational}, inheriting from the principle of Information Bottleneck \cite{tishby99information}, aims to learn fair representations that contain the minimal sufficient information related to the task.}
As a solution for the trade-off between utility and fairness in fair representation learning, the CFB is proposed to learn the conditional distribution $\rm P(\mathbf{Z}|\mathbf{X})$. This distribution captures the relationship between the input data $\mathbf{X}$ and the learned representation $\mathbf{Z}$, aiming to maintain sufficient task-relevant information not shared by sensitive attributes, while minimizing other task-irrelevant information. In this way, we can get fair representations that are informative of the task but contain little sensitive information. When encoding the representation $\mathbf{Z}$, the CFB aims to keep a certain level $r$ of the information relevant to the fair representation and downstream task $\mathbf{Y}$.
\begin{equation}
\resizebox{.9\hsize}{!}{$\min\limits_{\rm P(\mathbf{Z}|\mathbf{X})}\{I(\mathbf{S};\mathbf{Z})+I(\mathbf{X};\mathbf{Z}|\mathbf{S},\mathbf{Y})\}, \quad\mbox{s.t.} \quad I(\mathbf{Y};\mathbf{Z}|\mathbf{S})\geq r.$}
\end{equation}
The first term restricts the information that contains sensitive attributes $\mathbf{S}$ in the representation $\mathbf{Z}$, \oy{while the second term minimizes the information keeping in $\mathbf{Z}$ about the sensitive attributes $\mathbf{S}$ and the information about the data $\mathbf{X}$ irrelevant to the task $\mathbf{Y}$.} 
The constraint condition maintains the information relevant to the task $\mathbf{Y}$ not shared by sensitive attribute $\mathbf{S}$.


\subsection{Fairness definitions}
We will present a comprehensive overview of several widely recognized fairness definitions, and we focus on a common scenario for the binary sensitive attribute $s \in \{0,1\}$ and the binary label $y \in \{0,1\}$. The predictive label of the node classifier is denoted as $ \hat{y} \in \{0,1\}$. These fairness definitions provide various approaches to measuring biases.

\begin{definition}
\textbf{Statistical parity (Demographic parity)} \cite{wan2021modeling}. Statistical parity aims to ensure that the predictions are consistent across different demographic groups, particularly for sensitive attributes. Specifically, it means that groups with different sensitive attributes should have the same probability of positive prediction, which can be formally written by:
\begin{equation}
\mathbb P(\hat{y}=1|s=0) = \mathbb P(\hat{y}=1|s=1).
\end{equation}
\end{definition}
\begin{definition}
\textbf{Equal opportunity} \cite{dong2022fairness}. Equal opportunity aims to ensure that instances in a positive class from different demographic groups have an equal probability of receiving positive outcomes. To be more specific, equal opportunity states that different groups should have equal true positive rates, which is defined as:
\begin{equation}
\mathbb P(\hat{y}=1|s=0, y=1) = \mathbb P(\hat{y}=1|s=1, y=1).
\end{equation}
\end{definition}
\begin{definition}
\textbf{Counterfactual fairness}\cite{ma2022learning}. Counterfactual fairness aims to evaluate whether a model's output would remain the same even if the sensitive attribute of an individual was different. In other words, it evaluates the stability of a decision to the sensitive attribute value changes in the input data. Specifically, counterfactual fairness requires the prediction should be independent of the sensitive attribute. Suppose there exists a graph encoder $\Phi(\cdot)$, graph counterfactual fairness is defined as follows:
\begin{equation}
\resizebox{.9\hsize}{!}{$\mathbb P((\mathbf{z}_{i})_{s\leftarrow 0}|X=x, S=s)=\mathbb P((\mathbf{z}_{i})_{s\leftarrow 1}|X=x,S=s),$}
\end{equation}

where $\mathbf{z}_{i}=\Phi(\mathbf{X},\mathbf{A})_i$ denotes the learned representation of node $v_{i}$.
\end{definition}

\subsection{Problem definition} 

Based on the background described previously, we aim to develop a methodology for learning a fair graph encoder, a model designed to map node features represented by $\mathbf{X}$ to low dimensional node representations embedded within $\mathbf{Z}\in \mathbb{R}^{n \times d'}$. These learned representations are important in downstream graph tasks including node classification, link prediction, and graph classification.  In this paper, we only consider node classification tasks, but it is worth noting that our work can also be extended to link prediction and graph classification tasks. 

Typically, the sensitive attribute of a node $s_i$ is usually a binary variable following existing work \cite{dai2021say,wang2022improving,agarwal2021towards}. Let $y_i$ represents the true label of a node $v_i$, a node classifier takes the representation $\mathbf{z}_i\in \mathbf{Z}$ of the given node $v_i$ as input and outputs a predicted label $\hat{y}_i$. The goal of fair representation learning is to ensure that \mr{$\hat{\mathbf{Y}}=\{\hat{y}_1, \hat{y}_2, ..., \hat{y}_n\}$ achieves high accuracy while satisfying the fairness criteria.}



\section{Our Framework: GRAFair}
\label{method}

In this section, we propose a novel framework GRAFair which aims to learn fair node representations on graphs. 

\begin{figure*}[!t]
\centering
\includegraphics[width=6in]{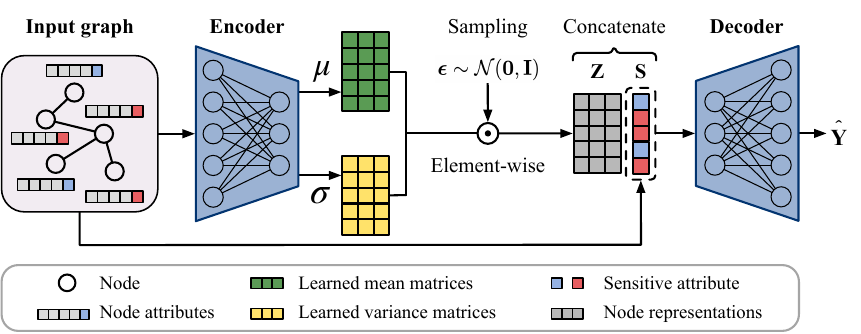}
\caption{An illustration of the proposed framework. GRAFair consists of two parts: an encoder and a decoder. The \oy{variational graph encoder maps the input graph data $\mathcal{G}$ to node representations $\mathbf Z$. The encoder learns the mean $\bm{\mu}_i$ and log variance $\log{\bm{\sigma}_i}$ of $\mathbf z_i$. By sampling $\bm \epsilon$ from standard Gaussian distribution, we can obtain the latent representation of the node $\mathbf z_i=\bm\mu_i+\bm\sigma_i \odot \bm\epsilon$.} The node representation $\mathbf Z$ sampling from the learned distribution together with sensitive attributes $\mathbf S$ are the input of the decoder during training. \oy{The decoder utilizes representations to predict label $\hat{\mathbf Y}$ in downstream tasks.}} 



\label{fig_2}
\end{figure*}

\subsection{Objective function}
\label{sec:GFIB}
We formulate our graph fair representation learning problem based on a variational graph auto-encoder (VGAE). As shown in Figure \ref{fig_2}, our framework GRAFair consists of two parts, an encoder mapping the original graph data $\mathcal{G}$ into a representation $\mathbf{Z}$ and a decoder performing node classification tasks based on the learned representation $\mathbf{Z}$. The ideal case is to make the learned representation $\mathbf{Z}$ from conditional likelihood $\rm P(\mathbf{Z}|\mathcal{G})$ independent of the sensitive attributes $\mathbf{S}$, which is clearly tricky to optimize for such a strong constraint. Naturally, the goal of fairness in representation learning becomes to minimize the mutual information $I(\mathbf{S};\mathbf{Z})$. In order to maintain the task-related information in the learned representations, we consider the Information Bottleneck item as a constraint. When placing alongside the above two considerations, we can formulate our optimization object following the CFB. The optimization problem can be converted into minimizing the Lagrangian of the CFB problem for learning a distribution $\rm P(\mathbf{Z}|\mathbf{G})$:
\begin{equation}
\label{eqn:lagrangian}
\min\limits_{\rm P(\mathbf{Z}|\mathcal{G})} I(\mathbf{S};\mathbf{Z})+I(\mathcal{G};\mathbf{Z}|\mathbf{S},\mathbf{Y})-\alpha I(\mathbf{Y};\mathbf{Z}|\mathbf{S})\text{,}
\end{equation}

\oy{where the first term minimizes the mutual information between sensitive attributes $\mathbf S$ and learned representations $\mathbf Z$, the second term minimizes the information related to sensitive attributes $\mathbf S$ and task-irrelated information preserved in representations $\mathbf Z$, and the third term maximizes the task-related information of the representation $\mathbf Z$.}


In a realistic scenario, credit institutions use the graph-structured data $\mathcal G$ of clients to determine whether to approve loan applications.
To mitigate discrimination in GNN predictions, we map $\mathcal G$ into representation $\mathbf Z$, which contains less sensitive information. The Markov chains $\mathbf S\leftrightarrow \mathcal G \rightarrow \mathbf Z$ and $\mathbf Y \leftrightarrow \mathcal G \rightarrow \mathbf Z$ hold throughout the process. Using Mutual Information properties and Markov chains, we can then write the following inspired by Graph Information Bottleneck \cite{wu2020graph}:

\begin{flalign}
  &\min\limits_{\rm P(\mathbf Z|\mathcal G)} I(\mathbf S;\mathbf Z)+I(\mathcal G;\mathbf Z|\mathbf S,\mathbf Y)-\alpha I(\mathbf Y;\mathbf Z|\mathbf S) \label{enq:7}\\
 =&\min\limits_{\rm P(\mathbf Z|\mathcal G)} I(\mathcal G;\mathbf Z)-I(\mathbf Y;\mathbf Z|\mathbf S)-\alpha I(\mathbf Y;\mathbf Z|\mathbf S)\label{enq:8} \\
 =&\min\limits_{\rm P(\mathbf Z|\mathcal G)} I(\mathcal G;\mathbf Z)-(\alpha+1) I(\mathbf Y;\mathbf Z|\mathbf S) \\
 =&\min\limits_{\rm P(\mathbf Z|\mathcal G)} I(\mathcal G;\mathbf Z)-\beta I(\mathbf Y;\mathbf Z|\mathbf S) \text{, where } \beta = \alpha + 1. \label{enq:betaloss}
\end{flalign}
Overall, our objective goal can be written in the following form:
\begin{equation}
\label{eqn:objectivegoal}
\min_{\rm P(\mathbf Z|\mathcal G)} I(\mathcal G;\mathbf Z)-\beta I(\mathbf Y;\mathbf Z|\mathbf S).
\end{equation}


\subsection{The solution to GRAFair}
\label{section:solution}

Based on the formula (\ref{eqn:objectivegoal}), we require the learned node representation $\mathbf Z$ to minimize the information from the graph dataset $\mathcal G$ and maximize the prediction $\mathbf Y$ under the sensitive attributes $\mathbf S$. However, $I(\mathcal G;\mathbf Z)$ and $I(\mathbf Y;\mathbf Z|\mathbf S)$ are still intractable for an exact optimization of $\rm P(\mathbf Z|\mathcal G)$. Therefore, we apply the variational approach, which is widely used for the optimization problem, to derive variational bounds of these two terms, solving intractable computation.

The term $I(\mathcal G;\mathbf Z)$ depends on the probabilistic distribution $\rm P(\mathbf Z|\mathcal G)$, and the term $I(\mathbf Y;\mathbf Z|\mathbf S)$ depends on the probabilistic distribution $\rm P(\mathbf Y,\mathbf Z|\mathbf S)$.
Due to the non-negativity of the relative entropy, we can bound the terms above based on previous works \cite{kingma2013auto,alemi2016deep}. For any probabilistic distribution $\rm P_\theta(\mathbf Z|\mathcal G)$ with parameter $\theta$ and $\rm Q(\mathbf Z)$, we have the upper bound of $I(\mathcal G;\mathbf Z)$ as follows:
\begin{equation}
\label{eqn:upperbound}
I(\mathcal G;\mathbf Z)\leq D_\text{KL}({\rm P}_\theta(\mathbf Z|\mathcal G)\|{\rm Q}(\mathbf Z)).
\end{equation}

For any probabilistic distribution $\rm P_\phi(\mathbf Y,\mathbf Z|\mathbf S)$ with parameter $\phi$ and $\rm Q(\mathbf Y|\mathbf S)$, we derive the lower bound of $I(\mathbf Y;\mathbf Z|\mathbf S)$ as follows (the proof details are in the Appendix):
\begin{equation}
\label{eqn:lowerbound}
I(\mathbf Y;\mathbf Z|\mathbf S)\geq \mathbb{E}_{\rm P(\mathbf Y,\mathbf Z, \mathbf S)}\left[\log\frac{\rm P_\phi(\mathbf Y|\mathbf Z,\mathbf S)}{\rm Q(\mathbf Y|\mathbf S)}\right].
\end{equation}

Then we can get the following loss function and jointly learn the parameters $\theta$ and $\phi$:
\begin{equation}
\begin{aligned}
\label{eqn:lossfunction}
\mathcal{L}= & D_\text{KL}(\rm P_\theta(\mathbf Z|\mathcal G)\|\rm Q(\mathbf Z))\\
&\qquad\qquad\quad-\beta \mathbb{E}_{\rm P(\mathbf Y,\mathbf Z, \mathbf S)}\left[\log\frac{\rm P_\phi(\mathbf Y|\mathbf Z,\mathbf S)}{\rm Q(\mathbf Y|\mathbf S)}\right].
\end{aligned}
\end{equation}
A posterior interpretation of the above approach from a coding viewpoint is that we use $I(\mathcal G;\mathbf Z)$ in Equation (\ref{enq:betaloss}) to encourage $\rm P_\theta(\mathbf Z|\mathcal G)$ to approach its marginal $\rm Q(\mathbf Z)$. \oy{At the same time, the encoder will generate the representation $Z$ while limiting the sensitive information contained in the original graph $G$}.

\oy{For the first term of formula (\ref{eqn:lossfunction}), suppose the true posterior distribution conforms to a Gaussian Mixture distribution ${\rm P}_\theta(\mathbf Z|\mathcal G)=\prod_i p_\theta(\mathbf z_i|\mathcal G)=\prod_i\mathcal{N}(\mathbf z_i|\bm \mu_i,\mathrm{diag}(\bm\sigma_i^2))$ and the approximate prior distribution ${\rm Q}(\mathbf Z)=\prod_i q(\mathbf z_i)=\prod_i \mathcal{N}(\mathbf z_i|\mathbf 0,\mathbf I)$.} To use gradient descent optimization techniques to learn the parameter $\theta$, we adopt the reparametrization trick \cite{kingma2013auto} to make the gradients calculable. The reparameterization trick is as follows:
\begin{equation}
\label{eqn:reparamterization}
\mathbf z_i=\bm\mu_i+\bm\sigma_i \odot \bm\epsilon,\quad \bm\epsilon\sim \mathcal{N}(\mathbf 0,\mathbf I),
\end{equation}
where the random variable $\mathbf z_i$ is transformed in a differentiable way, and $\bm\epsilon$ is an auxiliary variable sampling from standard Gaussian distribution $\mathcal{N}(\mathbf 0,\mathbf I)$, and $\odot$ denotes element-wise product. In this way, we can sample $\mathbf Z$ from the distribution above. 

 
 Existing works on Information Bottleneck (IB) only consider i.i.d data, such as tabular or image data, which can not be simply applied to graph data. 
 \oy{Inheriting from the principle of IB, we require node representation $\mathbf Z$ to minimize the undesirable information and maximize the information for the prediction task on graphs. To learn IB-based GRAFair, the model needs sample data points to derive variational bounds and accurately estimate those bounds \cite{alemi2016deep}.}
 \oy{However,} we can not sample a node in a connected graph directly while fully capturing the correlation in the underlying graph structure. \oy{In order to define a more tractable search space of the optimal $\rm P(\mathbf Z|\mathcal G)$ in graph-structure data}, we have to make some additional assumptions. We leverage a widely accepted local-dependence assumption \cite{wu2020graph} to make searching optimal distribution more tractable. The node $v_{i}$ in the graph will only be influenced by its neighbors within a certain number of hops, assuming the rest of the data is independent of node $v_{i}$. Based on this assumption, $\rm P(\mathbf Z|\mathcal G)$ and $\rm Q(\mathbf Z)$ can be written as ${\rm P}(\mathbf Z|\mathcal G)=\prod_{i=1}^{n}p(\mathbf z_i|G)$ and ${\rm Q}(\mathbf Z)=\prod_{i=1}^{n}q(\mathbf z_i)$.

For the second term of formula (\ref{eqn:lossfunction}), the estimation of $\rm Q(\mathbf Y|\mathbf S)$ can be derived from the empirical density of the data. So $I(\mathbf Y;\mathbf Z|\mathbf S)$ only depends on $\rm P_\phi(\mathbf Y,\mathbf Z|\mathbf S)$. To obtain the probability distribution of $\mathbf Y$ under multiple conditions, we concatenate the sensitive attribute $\mathbf S$ and the representation $\mathbf Z$ in latent space. Similarly, we have ${\rm P}_\phi(\mathbf Y|\mathbf Z,\mathbf S) = \prod_{i=1}^{n}p(y_i|\mathbf z_i\|s_i)$ under the local-dependence assumption, where the symbol $\|$ is the concatenation operator.

Finally, we can obtain the following loss function:
\begin{equation}
\begin{aligned}
\label{eqn:costfunction}
\mathcal{L}=&\frac{1}{N}\sum_{i=1}^{N}D_\text{KL}(p_\theta(\mathbf z_i|\mathcal G)\|q(\mathbf z_i))\\
&\qquad\qquad\quad-\beta \mathbb{E}_{{\rm P}(\mathbf Y,\mathbf Z, \mathbf S)}\left[\log p_\phi(y_i|\mathbf{z}_i||s_i)\right].
\end{aligned}
\end{equation}

\subsection{Training of GRAFair}
\label{section:algorithm}

\begin{algorithm}[!t]
        \label{alg:alg1}
        \caption{\oy{The algorithm of GRAFair.}}
        \KwIn{The graph dataset $\mathcal G$; Sensitive attributes $\mathbf S$; The number of training epochs $T$; Trade-off parameter $\beta$; Node set $\mathcal{V}$; Encoder layers $L$.}
        \KwOut{Fair representations $\mathbf Z_X^{(L)}$; Predictions $\hat{\mathbf Y}_v$.}
         
        \textbf{Initialize:} $\mathbf Z^{(0)}_X\leftarrow \mathbf X$; Encoder weights $\mathbf W^{(l)} \in \mathbb{R}^{f^{\prime} \times 2 f^{\prime}}$; Decoder weights $\mathbf W_{\text {out }} \in \mathbb{R}^{{(f^{\prime} +2)}\times K}$;\\
        \For{epoch$ \leftarrow 1, ..., T$}{
            \For{$l\leftarrow 1, ..., L \text{ and } v\in\mathcal{V}$}
            {
                    $\widetilde{\mathbf Z}^{(l)}_{X,v} \leftarrow \mathbf Z^{(l-1)}_{X,v}\mathbf W^{(l)}$; \\
                    $\mathbf Z_{A, v}^{(l)} \leftarrow \text{NeighborSample}(\mathbf Z^{l-1}_{X},\mathcal{V})$; (See \cite{wu2020graph})\\
                    $\bar{\mathbf Z}_{X, v}^{(l)} \leftarrow \sum_{u \in \mathbf Z_{A, v}^{(l)}} \tilde{\mathbf Z}_{X, v}^{(l-1)}$;\\
                    $\bm\mu_v^{(l)} \leftarrow \bar{\mathbf Z}_{X, v}^{(l)}\left[0: f^{\prime}\right]$;\\
                    $\bm\sigma_v^{2(l)} \leftarrow \operatorname{softplus}\left(\bar{\mathbf Z}_{X, v}^{(l)}\left[f^{\prime}: 2 f^{\prime}\right]\right)$; \\
                    Sample ${\mathbf Z}^{(l)}_{X,v}\sim\text{Gaussian}({\bm\mu}^{(l)}_v, {{\bm\sigma}_v^{2(l)}})$ ;\\
            }
            {
                $\hat{\mathbf Y}_v=\operatorname{softmax}\left((\mathbf Z_{X, v}\|\mathbf S) \mathbf W_{\text {out }}\right)$;\\
                $\textbf{update }\theta_E,\theta_D $ according to loss function; (See Eqn. (\ref{eqn:costfunction}))\\
            }
    
    \textbf{return}  ${\mathbf Z}^{(L)}_X$, $\hat{\mathbf Y}_v$
    }
    
\end{algorithm}

\oy{During the encoding process, we adopt the neighbor sampling method from Graph Information Bottleneck \cite{wu2020graph} for neighbor aggregation. The encoder learns the mean $\bm\mu_i$ and log variance $\log{\bm\sigma_i}$ of $\mathbf z_i$, then we can obtain the latent representation by sampling $\mathbf z_i$ from the Gaussian distribution $\mathcal{N}(\bm\mu_i,\bm\sigma^2_i)$. Before feeding the decoder for downstream tasks, we merge the sensitive attribute $\mathbf S$ with the sampled representation $\mathbf Z$. This concatenation can emphasize the ability of the decoder to capture sensitive information while weakening the ability of the encoder during training. Consequently, the well-trained encoder typically learns general task-relevant information rather than sensitive information.} At the application stage, the fair representation $\mathbf Z$ from the trained encoder does not need to splice sensitive attributes $\mathbf S$ and can perform downstream tasks individually.
The pseudo-code of the complete GRAFair framework is summarized in Algorithm \ref{alg:alg1}.


\section{Experiments}
\label{exp}
In this section, we conduct experiments on three real-world graph datasets to show the performance of our proposed framework GRAFair. We aim to answer the following questions: 

\textbf{Q1)} How does GRAFair perform compared to state-of-the-art baselines \oy{on utility}? 

\textbf{Q2)} \oy{How well does GRAFair promote fairness and stability?}

\textbf{Q3)} \oy{How does the time cost of our method compare with other baselines?}

\textbf{Q4)} How does the hyper-parameter $\beta$ influence the performance?

\textbf{Q5)} How do the components in GRAFair contribute to the performance?

\subsection{Experimental setup}

\subsubsection{Datasets} We perform experiments on three public real-world graph datasets. The detailed statistics of these datasets are shown in Appendix.

\textbf{German Credit Dataset (German)}. This dataset has 1,000 nodes, where each node represents a person who takes credit from a bank. Each node contains 27 attributes. The edge between the two nodes indicates that persons have similar credit behaviors. Each person is classified as having high or low credit risk according to their attributes. Gender is treated as a sensitive attribute. 

\textbf{Credit Default Dataset (Credit)}. This dataset has 30,000 nodes, where each node represents a person who uses credit cards. Each node contains 13 attributes. The edge between the two nodes indicates that persons have similar payment behaviors. According to their attributes, each person is classified as to whether they will default on their loans. Age is treated as a sensitive attribute. 

\textbf{Recidivism Dataset (Bail)}. This dataset has 18,876 nodes, where each node represents a criminal defendant. Each node contains 18 attributes. The edge between the two nodes indicates that persons have similar criminal behaviors. Each person is classified based on whether they will receive bail according to their attributes. Race is treated as a sensitive attribute. 

\begin{table*}[!t]
\renewcommand\arraystretch{1.2}  
\small
\caption{The performance (mean ± standard deviation over five repeated executions) of GRAFair based on GCN and other baselines. $\uparrow$ means larger is better, while $\downarrow$ means lower is better. (Bold: the best; underline: the runner-up.)\label{tab:table2}} 
\centering
\setlength\tabcolsep{10pt}

\begin{tabular}{l|l|lllll}
\hline
Datasets                & Baseline & F1-score($\uparrow$) & $ \Delta_{SP}(\downarrow) $ & $  \Delta_{EO}(\downarrow) $  & $  \Delta_{CF}(\downarrow) $  & $\Delta_{RS}$($\downarrow$)               \\ \hline
\multirow{7}{*}{Bail}   & Vanilla  & 77.50±0.87           & 8.54±0.75          & 5.95±0.59          & 9.01±3.02          & 21.98±1.64         \\
                        & \mr{Vanilla w/o S} & \mr{77.50±0.75}          & \mr{8.50±0.57}          & \mr{5.95±0.35}          & \mr{9.01±9.16}          & \mr{21.98±2.69}         \\
                        
                        & FairGNN  & 78.14±0.94          & 6.51±0.77          & 4.51±1.10          & 2.74±2.12          & 14.36±1.86         \\
                        & NIFTY    & 69.22±0.63          & \underline{4.19±0.70}    & 3.94±0.83          & \underline{0.86±0.10}    & 6.99±1.22          \\
                        & FairVGNN & 80.05±0.62          & 6.15±0.47          & 4.43±0.99          & 8.25±5.90          & \underline{5.67±2.07}    \\
                        & Graphair & 75.80±3.87          & 5.59±0.85          & 1.98±1.24          & 42.58±0.14         & 42.36±0.27         \\
                        & FairGT   & \textbf{98.04±0.42} & 5.34±0.32          & \textbf{0.80±0.01} & 0.92±0.23          & 47.48±0.33         \\
                        & GRAFair  & \underline{92.10±0.56}    & \textbf{1.18±0.27} & \underline{1.67±1.12}    & \textbf{0.00±0.00} & \textbf{3.78±0.38} \\ \hline
\multirow{7}{*}{Credit} & Vanilla  & 80.07±3.53          & 7.49±5.66          & 6.91±5.59          & 14.79±7.14         & 22.66±8.68         \\
                         & \mr{Vanilla w/o S} & \mr{78.50±0.13}          & \mr{8.75±31.97}         & \mr{8.27±29.34}         & \mr{17.88±3.97}        & \mr{26.10±21.82}        \\
                         
                        & FairGNN  & 76.72±1.81          & 13.50±5.01         & 12.86±5.35         & 18.52±12.36        & 15.03±3.46         \\
                        & FairVGNN & \underline{87.61±0.16}    & \underline{2.27±2.48}    & \underline{1.15±1.26}    & 3.98±2.16          & \underline{1.89±1.49}    \\
                        & NIFTY    & 79.96±0.06          & 9.76±0.14          & 8.59±0.28          & \underline{0.10±0.05}    & 6.82±1.07          \\
                        & Graphair & 78.96±11.73         & 12.08±9.54         & 12.39±13.91        & 37.49±4.38         & 38.89±5.78         \\
                        & FairGT   & 86.99±0.28          & 3.13±11.42         & 1.99±3.28          & 1.52±0.86          & 41.18±19.97        \\
                        & GRAFair  & \textbf{87.81±0.14} & \textbf{1.18±0.81} & \textbf{0.41±0.26} & \textbf{0.06±0.08} & \textbf{0.94±0.17} \\ \hline
\multirow{7}{*}{German} & Vanilla  & 79.00±3.20          & 43.18±4.36         & 32.79±5.18         & 24.04±4.50         & 12.00±1.02         \\
                         & \mr{Vanilla w/o S} & \mr{79.70±3.92}          & \mr{41.94±33.13}        & \mr{31.16±19.22}        & \mr{21.92±36.27}        & \mr{11.76±3.33}         \\
                        & FairGNN  & 81.82±0.32          & 38.33±5.02         & 27.58±4.65         & 14.56± 5.44        & 4.00±1.17          \\
                        & FairVGNN & \underline{82.45±0.17}    & \underline{1.44±2.57}    & \underline{0.92±1.10}    & 13.04±8.49         & 17.68±24.72        \\
                        & NIFTY    & 81.25±0.09          & 3.46±1.73          & 4.43±0.80          & \underline{0.48±0.44}    & \underline{0.72±0.76}    \\
                        & Graphair & 79.54±1.35          & 6.45±0.26          & 7.11±1.07          & 32.43±6.17         & 39.28±6.58         \\
                        & FairGT   & \textbf{84.08±1.22} & 3.19±4.71          & 4.47±4.64          & 5.76±5.01          & 11.20±3.28         \\
                        & GRAFair  & 80.95±0.00          & \textbf{0.81±0.47} & \textbf{0.78±0.56} & \textbf{0.27±0.14} & \textbf{0.68±0.55} \\ \hline
\end{tabular}
\end{table*}

\subsubsection{Evaluation metrics}The effectiveness evaluation of our proposed framework is from three aspects: classification performance, fairness, and robustness \cite{wang2023robust}. We use three fairness metrics (statistical parity, equal opportunity and counterfactual fairness) to evaluate fairness.

\textbf{F1-score.} We use the F1-score to measure the performance of classification tasks. The F1-score is a metric in binary classification that combines precision and recall into a single value, providing a balanced measure of a model's performance.

\textbf{Statistical Parity ($\Delta_{SP}$)}. Statistical parity denotes the equal distribution of positive outcomes among different demographic or sensitive groups, ensuring that the probability of receiving a positive prediction is consistent across these groups.
\begin{equation}
\Delta_{SP}=\left| \mathbb P(\hat{Y}=1|S=1)-\mathbb P(\hat{Y}=1|S=0)\right|
\end{equation}

\textbf{Equal Opportunity ($\Delta_{EO}$)}. Equal opportunity denotes that instances in a positive class should have an equal probability of being predicted to positive outcomes.
\begin{equation}
\Delta_{EO}=\left| \mathbb P(\hat{Y}=1|Y=1,S=1) -\mathbb P(\hat{Y}=1|Y=1,S=0)\right|
\end{equation}

\textbf{Counterfactual Fairness ($\Delta_{CF}$)}. Counterfactual fairness denotes that changing the sensitive attribute of an individual in a hypothetical scenario should not change the model's prediction or outcome for that individual.
\begin{equation}
\Delta_{CF}=\left| \mathbb P(\hat{Y}_{S\leftarrow 1}=Y|S=s)-\mathbb P(\hat{Y}_{S\leftarrow 0}=Y|S=s)\right|
\end{equation}

\textbf{Robustness score} ($\Delta_{RS}$). 
\oy{To assess the robustness of these models against noise (small perturbations to the node attributes), we take the percentage of label changes in the perturbed test nodes as the robustness score, following NIFTY \cite{agarwal2021towards}}. In our experiments, we draw a random attribute noise $\bm\delta\in {R_M}^{1\times d}$ sampled from a normal distribution. The node of perturbed attributes $n_{i}$ is then defined as $\widetilde{\mathbf x}_i =\mathbf x_i + \bm\delta$ (except for sensitive attributes). 

\begin{equation}
\Delta_{RS}=\left| \mathbb P(\hat{Y}_{X\leftarrow x}=Y|X=x)-\mathbb P(\hat{Y}_{X\leftarrow \widetilde{x}}=Y|X=x)\right|
\end{equation}

\subsubsection{Baselines}We compare GRAFair with five state-of-the-art \oy{fairness-aware methods}, \textit{i}.\textit{e}., FairGNN \cite{dai2021say}, NIFTY \cite{agarwal2021towards}, FairVGNN \cite{wang2022improving}, \oy{Graphair \cite{ling2022learning} and FairGT \cite{luo2024fairgt}}. Among them, FairGNN, FairVGNN and Graphair are adversarial representation learning methods, and NIFTY belongs to the filtering-based method. FairGT is a fairness-aware method for the graph transformer. \mr{Additionally, we include GCN \cite{kipf2016semi}, GIN \cite{xu2018powerful}, GraphSAGE \cite{hamilton2017inductive} and  Cheb \cite{defferrard2016convolutional} separately as vanilla baselines. These models represent the original architectures without any fairness-specific modifications. We also implement a straightforward debiasing approach, Vanilla w/o S, where sensitive attributes are removed from all nodes in the source dataset.}

\begin{figure*}[!t]
\small
\centering
\subfloat[Bail]{\includegraphics[width=2.2in]{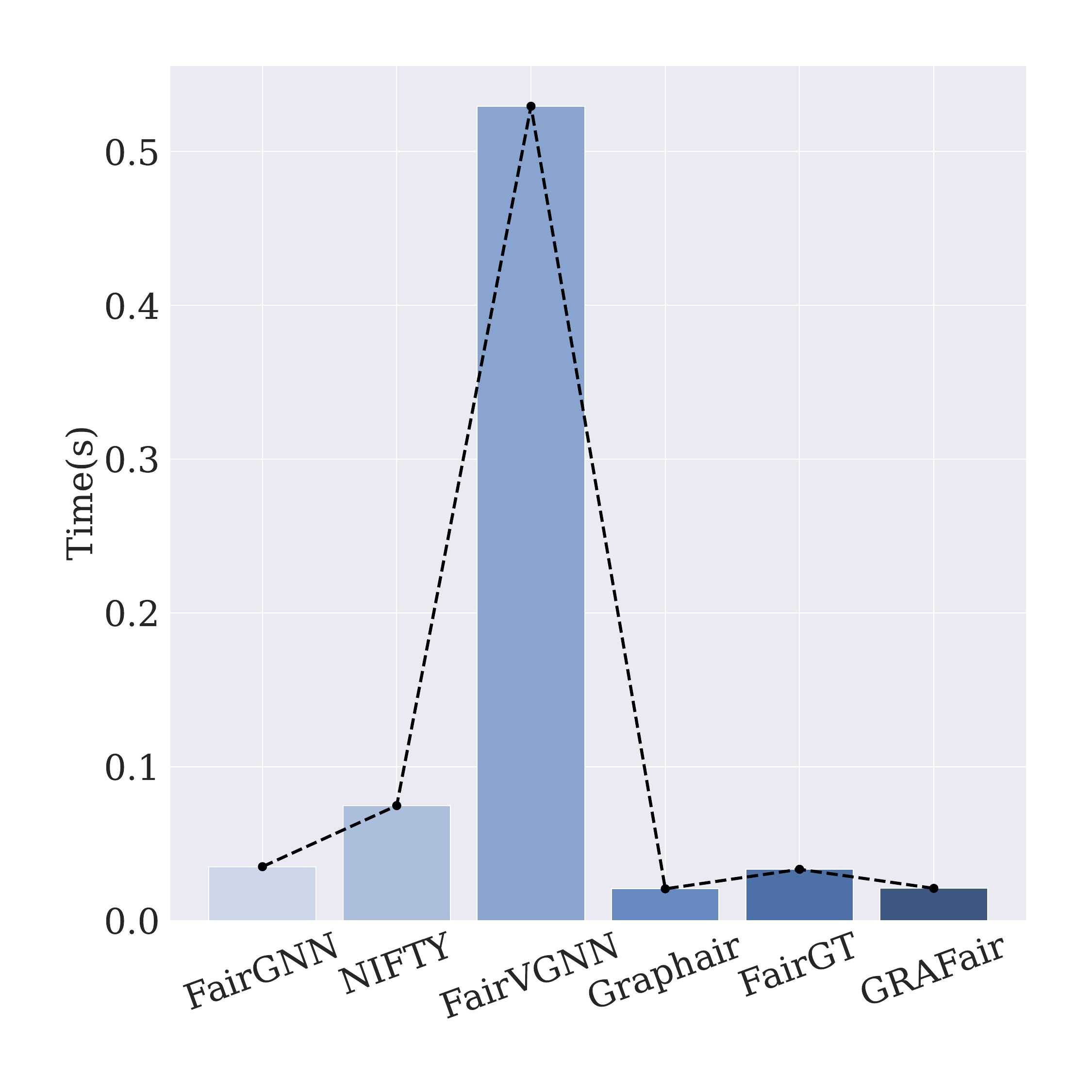}%
\label{fig_credit_time}}
\subfloat[Credit]{\includegraphics[width=2.2in]{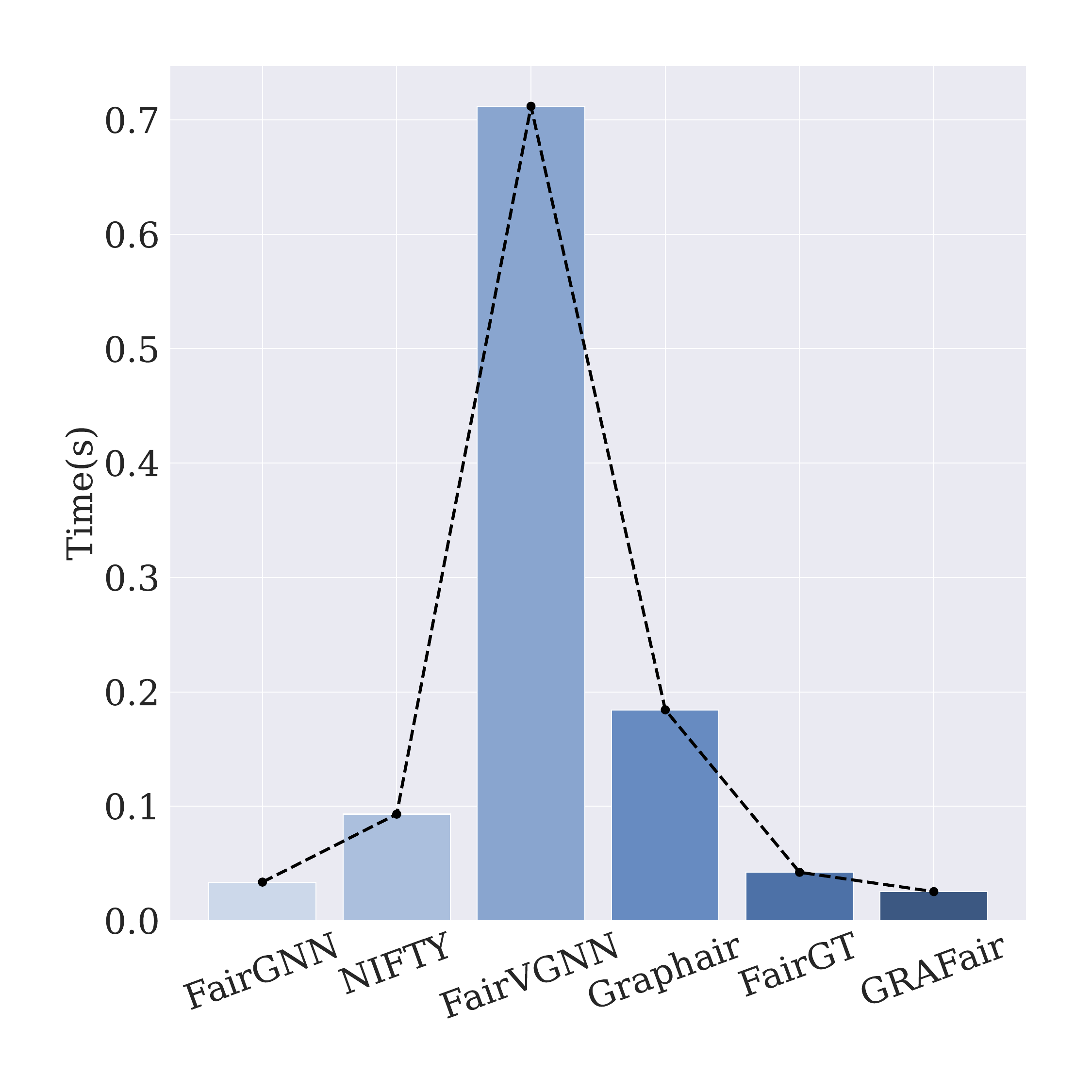}%
\label{fig_credit_time}}
\subfloat[German]{\includegraphics[width=2.2in]{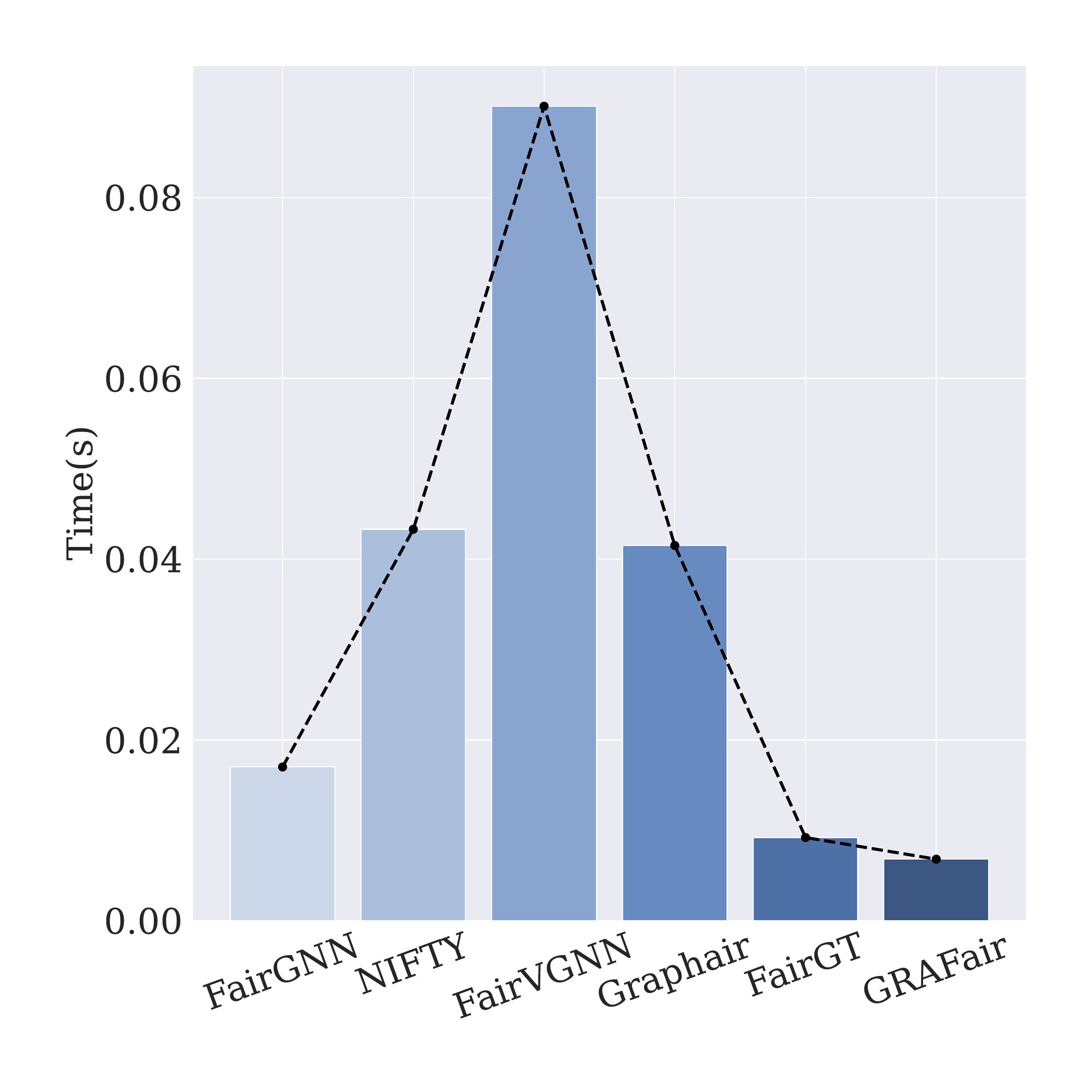}%
\label{fig_german_time}}
\caption{Time efficiency (in seconds) of different methods on Bail, Credit and German datasets. Each value refers to the average time during training of an epoch.}
\label{fig_time}
\end{figure*}
\textbf{FairGNN}. 
The FairGNN \cite{dai2021say} is a framework proposed to address discrimination in GNNs by learning fair representations with limited sensitive attribute information. It leverages graph structures and limited sensitive information to eliminate bias in GNNs while maintaining high node classification accuracy. The framework includes a GNN sensitive attribute estimator to predict sensitive attributes with noise for fair classification. An adversary is deployed to ensure the classifier makes predictions independent of the estimated sensitive attributes. Additionally, a fairness constraint is introduced to make the predictions invariant with the estimated sensitive attributes.

\textbf{NIFTY}. 
The NIFTY \cite{agarwal2021towards} framework is designed to enhance fairness and stability within GNNs through architectural and objective function modifications. The framework introduces graph augmentation and a triplet-based objective function to optimize the similarity between the original graph and its counterfactual and noisy representations.
NIFTY minimizes the difference in node representations between the original and augmented graphs to achieve fairness and robustness. The augmented graphs have counterfactual perturbations on the sensitive attributes or edges. 

\textbf{FairVGNN}. 
The FairVGNN \cite{wang2022improving} is designed to mitigate unfairness and discrimination in GNN predictions, particularly addressing the issue of sensitive attribute leakage during feature propagation in GNNs. The framework comprises two main modules: generative adversarial debiasing and adaptive weight clamping. \oy{The generative adversarial debiasing module aims to prevent sensitive attribute leakage from the input perspective by learning fair feature views. On the other hand, the adaptive weight clamping module aims to prevent sensitive attribute leakage from the model perspective by clamping weights of sensitive-correlated channels of the encoder.}


\textbf{Graphair}.
\oy{Graphair \cite{ling2022learning} is a pre-processing method to achieve fair graph representation learning via automated data augmentations. It trains an automated augmentation model based on adversarial learning, employing an adversary model to predict the sensitive attributes of nodes. This well-trained augmentation model generates new graphs with fair topology structures and node features while preserving the task-relevant information from the original graphs.}

\textbf{FairGT}.
\oy{FairGT \cite{luo2024fairgt} is a fairness-aware graph transformer, utilizing both structural topology and node feature encoding. In structural topology encoding, it employs eigenvectors corresponding to the $t$ largest magnitude eigenvalue of the adjacency matrix, ensuring a fairer representation of the structural topology. Meanwhile, in node feature encoding, FairGT considers k-hop information while preserving essential sensitive features for each node. This comprehensive approach enhances graph information encoding and ensures the independence of sensitive features, contributing to a fairness-aware training process.}

\subsubsection{Implementation} Considering that different GNN encoders may cause different degrees of unfairness, we conducted four representative GNN encoders in the experiments to evaluate the generality of our framework GRAFair: GCN \cite{kipf2016semi}, GIN \cite{xu2018powerful}, GraphSAGE \cite{hamilton2017inductive} and  Cheb \cite{defferrard2016convolutional}. 
\oy{We implement our model using PyTorch, and all experiments are run on a single GeForce GTX 3090 GPU with 24GB memory.}




\subsection{Results and discussion}
\subsubsection{Q1: Utility performance}
To validate the effectiveness of our proposed model, GRAFair, we conduct a comprehensive comparison with FairGT and other state-of-the-art baselines based on GCN. As shown in Table \ref{tab:table2}, the F1-score evaluations across three real-world datasets demonstrate the excellent performance of GRAFair in node classification tasks. GRAFair showcases a notable improvement in model utility, surpassing the vanilla GCN by approximately 11\%. This indicates the ability of GRAFair to mitigate undesirable influences stemming from the inherent bias of the datasets. 

\oy{FairGT outperforms all GNN-based methods on the Bail and German due to the powerful representation capabilities of graph transformers. However, GRAFair demonstrates comparable performance in comparison with other baselines, outperforming the leading GNN, FairVGNN, by around 15\% on the bail dataset. Furthermore, the fact that our framework maintains excellent performance across different GNN encoders reflects the generalizability of our framework.}



\begin{table*}[!t]
\small
\renewcommand\arraystretch{1.2}  
\caption{Ablation study on the Information Bottleneck item. It shows the performance (mean ± standard deviation over five repeated executions) of GRAFair based on GCN. $\uparrow$ means larger is better, while $\downarrow$ means lower is better. (Bold: the best.)\label{tab:table_ablation}}
\centering
\setlength\tabcolsep{10pt}
\begin{tabular}{l|l|lllll}
\hline
Datasets                & Baseline        & F1-score($\uparrow$) & $ \Delta_{SP}(\downarrow) $ & $  \Delta_{EO}(\downarrow) $  & $  \Delta_{CF}(\downarrow) $  & $\Delta_{RS}$($\downarrow$)        \\ \hline
\multirow{5}{*}{Bail}   & Vanilla         & 77.50±0.87           & 8.54±0.75          & 5.95±0.59          & 9.01±3.02          & 21.98±1.64         \\
                        & GRAFair$^\text{(-)}$  & 90.36±0.32          & 5.29±0.24          & \textbf{0.39±0.28} & 0.21±0.18          & 11.48±2.01         \\
                        & GRAFair$^\text{(\#)}$   & 92.02±0.07          & 13.12±0.07         & 1.55±0.12          & 0.10±0.13          & 3.88±0.19          \\
                        & \mr{GRAFair$^\text{(GAE)}$} & \mr{91.24±0.48}          & \mr{5.83±0.34}          & \mr{1.12±0.43}          & \mr{0.19±0.11}          & \mr{7.68±1.31}          \\
                        & GRAFair         & \textbf{92.10±0.56} & \textbf{1.18±0.27} & 1.67±1.12          & \textbf{0.06±0.02} & \textbf{3.78±0.38} \\ \hline
\multirow{5}{*}{Credit} & Vanilla         & 80.07±3.53          & 7.49±5.66          & 6.91±5.59          & 14.79±7.14         & 22.66±8.68         \\
                        & GRAFair$^\text{(-)}$  & 87.72±0.12          & 4.66±0.62          & 2.56±0.96          & 1.27±1.29          & 4.23±0.53          \\
                        & GRAFair$^\text{(\#)}$   & 86.30±4.42          & 1.62±3.28          & 1.39±3.06          & \textbf{0.03±0.05} & \textbf{0.07±0.10} \\
                        & \mr{GRAFair$^\text{(GAE)}$} & \mr{87.14±0.67}          & \mr{2.59±0.53}          & \mr{2.18±1.18}          & \mr{0.97±0.25}          & \mr{3.25±0.63}          \\
                        & GRAFair         & \textbf{87.81±0.14} & \textbf{1.18±0.81} & \textbf{0.41±0.26} & 0.06±0.08          & 0.94±0.17          \\ \hline
\multirow{5}{*}{German} & Vanilla         & 79.00±3.20          & 43.18±4.36         & 32.79±5.18         & 24.04±4.50         & 12.00±1.02         \\
                        & GRAFair$^\text{(-)}$  & 76.74±2.95          & 8.93±7.05          & 9.17±7.41          & 8.80±7.85          & 16.2±7.85          \\
                        & GRAFair$^\text{(\#)}$   & 78.05±0.00          & 1.83±0.62          & 1.20±0.84          & 0.81±0.52          & 1.40±1.13          \\
                        & \mr{GRAFair$^\text{(GAE)}$} & \mr{78.93±1.64}          & \mr{5.41±3.28}          & \mr{4.26±2.07}          & \mr{4.82±2.94}          & \mr{9.28±3.85}          \\
                        & GRAFair         & \textbf{80.95±0.00} & \textbf{0.81±0.47} & \textbf{0.78±0.56} & \textbf{0.27±0.14} & \textbf{0.68±0.55} \\ \hline
\end{tabular}
\end{table*}

\begin{figure}[!b]
\vspace{-0.4cm}
\centering
\setlength{\abovecaptionskip}{-0.2cm}
\includegraphics[width=3.6in]{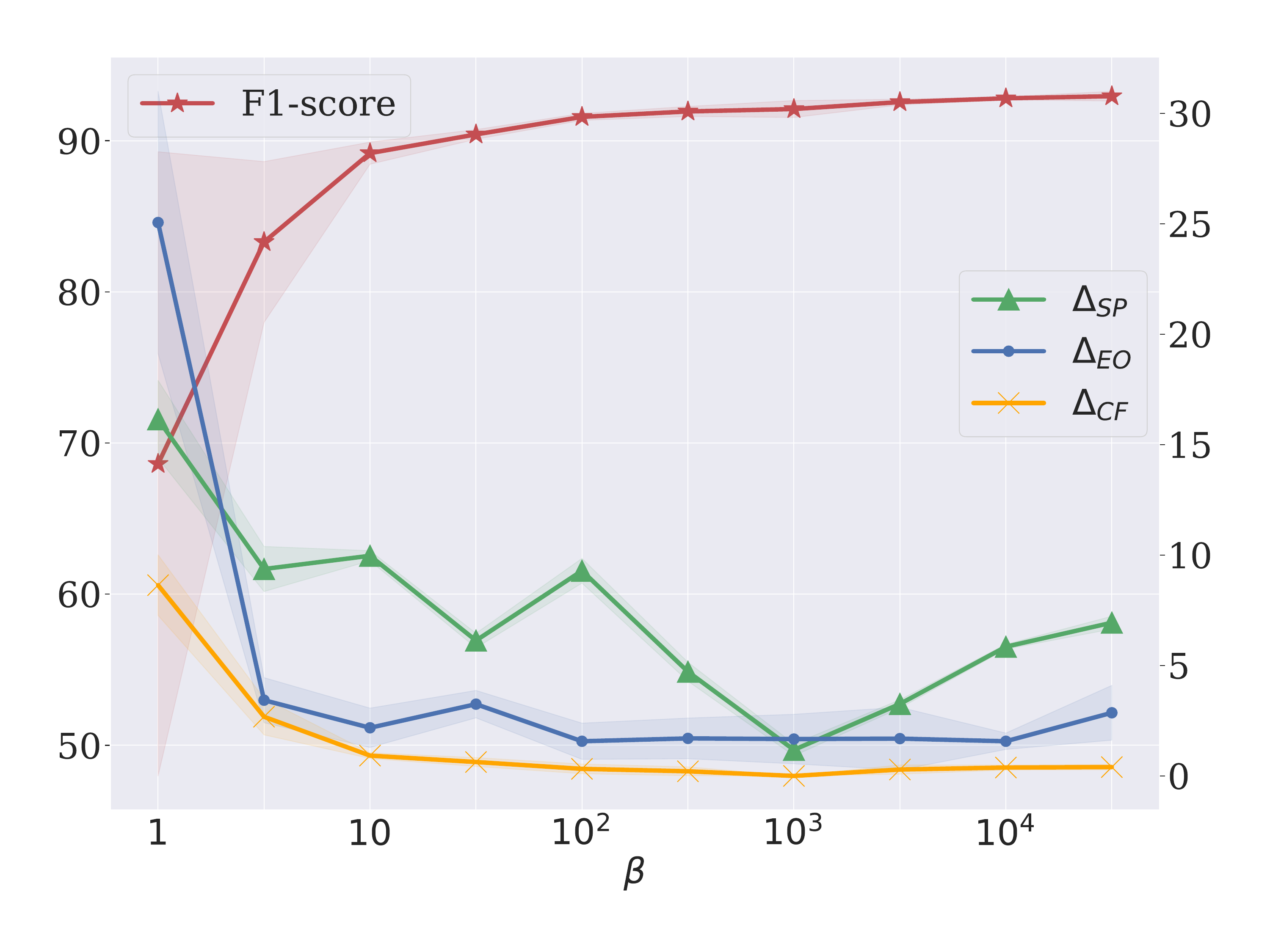}
\caption{Utility performance and fairness under different Hyper-parameter $\beta$ on Bail dataset. \mr{The value range of $\beta$ is $\{1,5,10,50,10^2,5\times 10^2,10^3,5\times 10^3, 10^4, 5\times 10^4\}$.} Here, $\beta=10^3$ can reach a favorable trade-off between utility and fairness.} 
\label{fig_4}
\end{figure}

\subsubsection{Q2: \oy{Debiasing performance and stability}}
\oy{To comprehensively demonstrate the debiasing performance of GRAFair and other baselines, we evaluate them on three widely used fairness metrics. As shown in Table \ref{tab:table2}, GRAFair is outstanding across three real-world datasets, proving the effectiveness of the proposed method in the fairness-aware node classification. In addition, observation can be drawn that GRAFair consistently exhibits the lowest variance across evaluation metrics, demonstrating the stability of our model.
Moreover, the robustness scores demonstrate that GRAFair outperforms other baselines in terms of robustness against noise perturbation. Thus our method is both efficient and stable, enabling potential applications in various scenarios.}

\mr{Besides, as shown in Table \ref{tab:table2}, Vanilla w/o S exhibits similar performance to Vanilla across various metrics on each dataset. This suggests that merely removing the sensitive attribute $\mathbf S$ from the node attribute $\mathbf X$ of the dataset does not significantly enhance the fairness of the model. This phenomenon arises due to the fact that sensitive information is not solely confined to the sensitive attributes, and the model can also capture implicit sensitive information from the non-sensitive node attributes and structural characteristics of the graph data\cite{mehrabi2021survey}.}

\subsubsection{Q3: Time efficiency}
\oy{As shown in Figure \ref{fig_time}, we present the training time cost of GRAFair with the baselines on Bail, Credit and German datasets. The lowest time cost among all methods demonstrates the efficiency of our method. The high time cost of FairVGNN is due to its adversarial training process and a large number of parameters \cite{wang2022improving}. In addition, we have included the time cost of different encoders among GNN methods in Table \ref{tab:table_encoder_time} in the Appendix. 
Furthermore, the time complexity of the encoder is determined by the GNN backbone used. In variational inference for the encoder, a multilayer perceptron (MLP) is commonly used, which is negligible compared to the GNN backbone. The decoder of GRAFair also uses a multilayer perceptron as the classifier. In short, GRAFair shares the same time complexity as other L-layer GNN backbones.}

\subsubsection{Q4: Hyper-parameter $\beta$ analysis} 


\oy{The hyper-parameter $\beta$ serves as a pivotal factor, representing the ratio between $I(\mathcal G;\mathbf Z)$ and $I(\mathbf Y;\mathbf Z|\mathbf S)$. 
To investigate the effect of hyper-parameter $\beta$, we experimented with various candidate $\beta$ over \mr{\{$1$, $5$, $10$, $10^2$, $5\times 10^2$, $10^3$, $5\times 10^3$, $10^4$, $5\times 10^4$\}}.
Figure \ref{fig_4} illustrates different trade-offs between utility and fairness. There is a clear trend that the utility performance of our model improves as $\beta$ increases. 
This trend can be attributed to the increased weight assigned to $I(\mathbf Y; \mathbf Z|\mathbf S)$, indicating a heightened focus on preserving the predictive performance of the model. It is important to choose a proper value of $\beta$, as setting it too low may lead to sensitive information leakage. 
Furthermore, simply increasing $\beta$ does not monotonously optimize fairness since sensitive information is controlled by both the term $I(\mathcal G; \mathbf Z)$ and the term $I(\mathbf Y; \mathbf Z|\mathbf S)$. So, we conducted experiments to find the optimal $\beta$ value to achieve a favorable trade-off between utility and fairness.}

\subsubsection{Q5: Ablation study} 

As formalized by the derived loss function in Equation (\ref{enq:betaloss}), our proposed framework endeavors to achieve fair representation learning through dual objectives: maximizing information about the target without sensitive information ($I(\mathbf Y;\mathbf Z|\mathbf S)$) and minimizing irrelevant-task information ($I(\mathcal G;\mathbf Z)$). We executed various ablation studies to elucidate the specific contributions of individual components within GRAFair to its performance.

Firstly, the impact of maximizing $I(\mathbf Y;\mathbf Z|\mathbf S)$ is evident when comparing GRAFair$^\text{(-)}$ with the vanilla model (original GCN) as shown in Table \ref{tab:table_ablation}. GRAFair$^\text{(-)}$ significantly enhances fairness and robustness while maintaining utility, indicating that maximizing $I(\mathbf Y;\mathbf Z|\mathbf S)$ weakens the ability of the model to capture sensitive information.

Secondly, we conduct an ablation on discouraging irrelevant-task information by $I(\mathcal G;\mathbf Z)$. For convenience, we denote GRAFair$^\text{(-)}$ as the model solely optimized by $I(\mathbf Y;\mathbf Z|\mathbf S)$. The results demonstrate that GRAFair consistently outperforms GRAFair$^\text{(-)}$ in most cases, both in terms of utility and fairness. This observation suggests that minimizing $I(\mathcal G;\mathbf Z)$ effectively reduces potentially sensitive information in representations derived from the original data.

Thirdly, to demonstrate the impact of integrating sensitive attributes $S$ into representation $Z$, we conduct the ablation studies experiments only optimized by $I(\mathcal G;\mathbf Z)-\beta I(\mathbf Y;\mathbf Z)$, denoted as GRAFair$^\text{(\#)}$. GRAFair performs better on fairness in Table \ref{tab:table_ablation}, which confirms our claim that concatenating $\mathbf S$ into $\mathbf Z$ weakened the ability of the encoder to capture the sensitive information.


\mr{Finally, we conduct an ablation experiment to demonstrate the effectiveness of VGAE compared with non-probabilistic graph auto-encoder (GAE) \cite{kipf2016variational} in Table \ref{tab:table_ablation}. GRAFair$^\text{(GAE)}$ indicates that GRAFair drops the variational part and uses a regular auto-encoder with the concatenation of $\mathbf S$ in the latent representation. 
By observing Table \ref{tab:table_ablation}, it can be noticed that GRAFair$^\text{(GAE)}$ performs poorly in both utility and debiasing compared to GRAFair. The first term in Equation (\ref{eqn:costfunction}) serves to limit the task-irrelevant information in $\mathbf Z$ in the form of variations, but this cannot be achieved using GAE alone.}

\subsection{Limitations and future works}

\oy{Though extensive experimental results demonstrate the effectiveness of GRAFair, the proposed method based on VGAE entails two common limitations in variational approaches to optimization. 
First, it estimates both decoding and marginal distributions that follow the restrictions of the variational approximation. This issue limits the search space of the possible encoding distributions, denoted as $\rm P(\mathbf Y|\mathbf X)$. 
Second, variational approaches heavily rely on parametrized densities. This issue further limits the search space of encoding distributions with densities $\rm P(\mathbf Y|\mathbf X, \theta)$, where $\theta$ represents the parameterization. To address these challenges, exploring richer encoding distributions and marginals offers a promising direction for alleviating these limitations, such as employing normalizing flows \cite{kingma2016improved}.}

\oy{Additionally, although the debiasing strategy for the single sensitive attribute of GRAFair has shown effective results across three datasets, some datasets may contain multiple sensitive attributes in real scenarios.
GRAFair cannot be directly applied to address multiple sensitive attributes because the interplay and trade-offs between these attributes can introduce new challenges.
Future research will broaden fairness considerations to encompass various forms of sensitive data. Additionally, efforts will focus on rectifying structural biases inherent in graph topology to enhance fairness across diverse real-world contexts.}

\section{Conclusion}
\label{conclusion}

This study concentrates on learning fair representations on graphs that can achieve fairness and maintain a good task-related performance simultaneously. More specifically, we aim to reduce sensitive information of interest from the learned representations in the training stage.
Inspired by the Conditional Fairness Bottleneck, we introduce GRAFair, a novel framework based on variational auto-encoder architecture. This method navigates the fairness-utility trade-off without relying on adversarial learning. To this end, GRAFair captures as much task-related information as possible while limiting sensitive features and task-irrelevant information from the graph.
Empirical evaluations on real-world datasets demonstrate the effectiveness of of GRAFair. It outperforms state-of-the-art baselines, exhibiting a superior fairness-utility trade-off alongside exceptional robustness, stability, and time efficiency.



\IEEEpubidadjcol



\newpage
{\appendices
\section{Detailed proof of equations (\ref{enq:7}) to (\ref{enq:8})}
\mr{First, we present the commonly used properties between conditional entropy, joint entropy, and mutual information:
\begin{equation*}
\begin{aligned}
\label{eqn:igz}
&I(\rm X; Y)=H(X)-H(X|Y)=H(Y)-H(Y|X),\\
&H(\rm Y|X)=H(X,Y)-H(X).\\
\end{aligned}
\end{equation*}
Based on the above fundamental properties, we derive $I(\mathbf S;\mathbf Z)+I(\mathcal G;\mathbf Z|\mathbf S,\mathbf Y)$ in Equation (\ref{enq:7}) as follows:
\begin{equation*}
\begin{aligned}
\label{eqn:igz}
&\min\limits_{\rm P(\mathbf Z|\mathcal G)} I(\mathbf S;\mathbf Z)+I(\mathcal G;\mathbf Z|\mathbf S,\mathbf Y)\\
 =&\min\limits_{\rm P(\mathbf Z|\mathcal G)} H(\mathbf S)+H(\mathbf Z)-H(\mathbf S,\mathbf Z)+H(\mathcal G|\mathbf S,\mathbf Y)\\
 &\qquad\qquad+H(\mathbf Z|\mathbf S,\mathbf Y)-H(\mathcal G,\mathbf Z|\mathbf S,\mathbf Y) \\
 =&\min\limits_{\rm P(\mathbf Z|\mathcal G)} H(\mathbf S)+H(\mathbf Z)-H(\mathbf S,\mathbf Z)+H(\mathcal G|\mathbf S,\mathbf Y)\\
 &\qquad\qquad+H(\mathbf Z|\mathbf S,\mathbf Y)-H(\mathcal G|\mathbf S,\mathbf Y)-H(\mathbf Z|\mathcal G,\mathbf S,\mathbf Y)\\
 =&\min\limits_{\rm P(\mathbf Z|\mathcal G)} H(\mathbf S)+H(\mathbf Z)-H(\mathbf S,\mathbf Z)+H(\mathbf Z,\mathbf S,\mathbf Y)\\
 &\qquad\qquad-H(\mathbf S,\mathbf Y)+H(\mathcal G,\mathbf S,\mathbf Y)-H(\mathcal G,\mathbf Z,\mathbf S,\mathbf Y)\\
 =&\min\limits_{\rm P(\mathbf Z|\mathcal G)} H(\mathcal G)+H(\mathbf Z)-H(\mathcal G,\mathbf Z)-\left[H(\mathbf S,\mathbf Y)-H(\mathbf S)\right]\\
 &\qquad\qquad-\left[H(\mathbf Z,\mathbf S)-H(\mathbf S)\right]+H(\mathbf Z,\mathbf S,\mathbf Y)-H(\mathbf S)\\
 =&\min\limits_{\rm P(\mathbf Z|\mathcal G)} H(\mathcal G)+H(\mathbf Z)-H(\mathcal G,\mathbf Z)-H(\mathbf Y|\mathbf S)-H(\mathbf Z|\mathbf S)\\
 &\qquad\qquad+H(\mathbf Y,\mathbf Z|\mathbf S)\\
 =&\min\limits_{\rm P(\mathbf Z|\mathcal G)} I(\mathcal G;\mathbf Z)-I(\mathbf Y;\mathbf Z|\mathbf S) \\
\end{aligned}
\end{equation*}
As shown above, Equation (\ref{enq:7}) can be derived into Equation (\ref{enq:8}):
\begin{equation*}
\begin{aligned}
  &\min\limits_{\rm P(\mathbf Z|\mathcal G)} I(\mathbf S;\mathbf Z)+I(\mathcal G;\mathbf Z|\mathbf S,\mathbf Y)-\alpha I(\mathbf Y;\mathbf Z|\mathbf S) \\
 =&\min\limits_{\rm P(\mathbf Z|\mathcal G)} I(\mathcal G;\mathbf Z)-I(\mathbf Y;\mathbf Z|\mathbf S)-\alpha I(\mathbf Y;\mathbf Z|\mathbf S) \\
 \end{aligned}
\end{equation*}
}


\section{Proof of the upper bound and lower bound}
\textbf{A1. The upper bound of $I(\mathcal G;\mathbf Z)$}.

 The upper bound of $I(\mathcal G;\mathbf Z)$ is derived from a variational approach \cite{agakov2004algorithm}. For any $\rm P(\mathbf Z|\mathcal G)$ and $\rm Q(\mathbf Z)$, we have:

\begin{equation*}
\begin{aligned}
\label{eqn:igz}
I(\mathcal G;\mathbf Z) &= \mathbb{E}_{{\rm P}(\mathcal G,\mathbf Z)} \left[ \log \frac{\rm P(\mathbf Z|\mathcal G)}{\rm P(\mathbf Z)}\right]\\
&=\mathbb{E}_{{\rm P}(\mathcal G,\mathbf Z)}\left[\log \frac{\rm P(\mathbf Z|\mathcal G)\rm Q(\mathbf Z)}{\rm P(\mathbf Z)\rm Q(\mathbf Z)}\right]\\
&=\mathbb{E}_{{\rm P}(\mathcal G,\mathbf Z)}\left[\log \frac{\rm P(\mathbf Z|\mathcal G)}{\rm Q(\mathbf Z)}\right] - \text{KL}(\rm P(\mathbf Z)||\rm Q(\mathbf Z))\\
&\leq \mathbb{E}_{{\rm P}(\mathcal G)}\left[\text{KL}(\rm P(\mathbf Z|\mathcal G)||\rm Q(\mathbf Z))\right].\\
\end{aligned}
\end{equation*}

\textbf{A2. The lower bound of $I(\mathbf Y;\mathbf Z|\mathbf S)$}.

The lower bound of $I(\mathbf Y;\mathbf Z|\mathbf S)$ is derived from \cite{kingma2013auto, kolchinsky2019nonlinear, alemi2016deep}. For any $\rm P(\mathbf Y|\mathbf Z,\mathbf S)$ and $\rm Q(\mathbf Y|\mathbf S)$, we have:

\begin{equation*}
\begin{aligned}
\label{eqn:igz}
I(\mathbf Y;\mathbf Z|\mathbf S)&= \int p(y,z|s)p(s)\mathrm{d}y\mathrm{d}z\mathrm{d}s\log \frac{p(y,z|s)}{p(y|s)p(z|s)}\\
&=\int p(y,z,s)\mathrm{d}y\mathrm{d}z\mathrm{d}s \log \frac{p(y|z,s)}{p(y|s)}\\
&\geq 1+ \mathbb{E}_{\rm P(\mathbf Y,\mathbf Z, \mathbf S)} \left[ \log \frac{\rm P(\mathbf Y|\mathbf Z,\mathbf S)}{\rm Q(\mathbf Y|\mathbf S)}\right ]\\
&\qquad\qquad\qquad+\mathbb{E}_{\rm P(\mathbf Y|\mathbf S)P(\mathbf Z)} \left[ \log \frac{\rm P(\mathbf Y|\mathbf Z,\mathbf S)}{\rm Q(\mathbf Y|\mathbf S)}\right ]\\
&\geq \mathbb{E}_{\rm P(\mathbf Y,\mathbf Z, \mathbf S)} \left[ \log \frac{\rm P(\mathbf Y|\mathbf Z,\mathbf S)}{\rm Q(\mathbf Y|\mathbf S)}\right ]\\
&=\mathbb{E}_{\rm P(\mathbf Y,\mathbf Z, \mathbf S)} \left[ \log \rm P(\mathbf Y|\mathbf Z,\mathbf S)\right ]\\
&\qquad\qquad\qquad-\mathbb{E}_{\rm P(\mathbf Y,\mathbf Z, \mathbf S)} \left[ \log \rm Q(\mathbf Y|\mathbf S)\right ].
\end{aligned}
\end{equation*}

where the Kullback Leibler divergence is always non-negative:
\begin{equation*}
\begin{aligned}
\label{eqn:kl}
&\text{KL}\left[ \rm P(\mathbf Y|\mathbf Z)||Q(\mathbf Y|\mathbf Z)\right] \geq 0 \Rightarrow \\ 
& \int  p(y|z)\log p(y|z) \mathrm{d}y\geq \int  p(y|z)\log q(y|z) \mathrm{d}y.
\end{aligned}
\end{equation*}

\section{The performance of fairness-aware GNNs based on different encoders}

The performance of fairness-aware GNNs based on different encoders is shown in Table \ref{tab:table_decoder}.

\begin{table*}[!t]
\renewcommand\arraystretch{1.2}  
\caption{The performance (mean ± standard deviation over five repeated executions) of fairness-aware GNNs based on different encoders: GCN, SAGE, Cheb, and GIN. $\uparrow$ means larger is better, while $\downarrow$ means lower is better. (Bold: the best.)\label{tab:table_decoder}}
\centering
\scriptsize
\setlength\tabcolsep{4.5pt}

\begin{tabular}{c|c|ccccc|ccccc}
\hline
\multirow{2}{*}{Datasets} & \multirow{2}{*}{Baseline} & \multicolumn{5}{c|}{GCN}                                                                                & \multicolumn{5}{c}{SAGE}                                                                                \\
                          &                           & F1-score($\uparrow$) & $ \Delta_{SP}(\downarrow) $ & $  \Delta_{EO}(\downarrow) $  & $  \Delta_{CF}(\downarrow) $  & $\Delta_{RS}$($\downarrow$)               & F1-score($\uparrow$) & $ \Delta_{SP}(\downarrow) $ & $  \Delta_{EO}(\downarrow) $  & $  \Delta_{CF}(\downarrow) $  & $\Delta_{RS}$($\downarrow$) \\ \hline
\multirow{5}{*}{Bail}     & Vanilla                   & 77.50±0.87           & 8.54±0.75          & 5.95±0.59          & 9.01±3.02          & 21.98±1.64         & 81.62±1.44          & 1.82±1.52          & 2.15±0.23          & 6.40±1.28          & 41.47±7.10         \\
                          & FairGNN                   & 78.14±0.94          & 6.51±0.77          & 4.51±1.10          & 2.74±2.12          & 14.36±1.86         & 81.30±0.66          & 2.03±1.20          & 1.25±1.17          & 9.40±1.78          & 24.74±2.15         \\
                          & FairVGNN                  & 80.05±0.62          & 6.15±0.47          & 4.43±0.99          & 8.25±5.90          & 5.67±2.07          & 84.46±0.75          & 3.15±1.39          & 1.97±1.16          & 24.01±21.27        & 15.99±12.52        \\
                          & NIFTY                     & 69.22±0.63          & 4.19±0.70          & 3.94±0.83          & 0.86±0.10          & 6.99±1.22          & 69.97±13.05         & 5.22±1.43          & 4.91±2.09          & 0.31±0.35          & \textbf{5.74±2.60} \\
                          & GRAFair                   & \textbf{92.10±0.56} & \textbf{1.18±0.27} & \textbf{1.67±1.12} & \textbf{0.00±0.00} & \textbf{3.78±0.38} & \textbf{99.33±0.17} & \textbf{1.48±0.06} & \textbf{0.29±0.19} & \textbf{0.04±0.02} & 9.39±0.75          \\ \hline
\multirow{5}{*}{Credit}   & Vanilla                   & 80.07±3.53          & 7.49±5.66          & 6.91±5.59          & 14.79±7.14         & 22.66±8.68         & 82.15±0.38          & 12.48±0.84         & 10.36±0.58         & 9.10±3.01          & 41.21±12.29        \\
                          & FairGNN                   & 76.72±1.81          & 13.50±5.01         & 12.86±5.35         & 18.52±12.36        & 15.03±3.46         & 79.53±1.29          & 11.18±1.06         & 9.36±0.74          & 23.78±13.31        & 30.72±5.74         \\
                          & FairVGNN                  & 87.61±0.16          & 2.27±2.48          & 1.15±1.26          & 3.98±2.16          & 1.89±1.49          & 87.32±1.01          & 8.46±5.00          & 5.60±3.68          & 18.17±15.81        & 10.09±7.03         \\
                          & NIFTY                     & 79.96±0.06          & 9.76±0.14          & 8.59±0.28          & 0.10±0.05          & 6.82±1.07          & 83.38±2.42          & 9.52±2.76          & 7.71±2.58          & 0.50±0.33          & 5.80±1.18          \\
                          & GRAFair                   & \textbf{87.81±0.14} & \textbf{1.18±0.81} & \textbf{0.41±0.26} & \textbf{0.06±0.08} & \textbf{0.94±0.17} & \textbf{87.61±0.07} & \textbf{0.13±0.3}  & \textbf{0.07±0.16} & \textbf{0.00±0.00} & \textbf{0.00±0.00} \\ \hline
\multirow{5}{*}{German}   & Vanilla                   & 79.00±3.20          & 43.18±4.36         & 32.79±5.18         & 24.04±4.50         & 12.00±1.02         & 80.43±1.05          & 25.95±5.30         & 17.69±6.58         & 9.52±5.11          & 6.72±2.78          \\
                          & FairGNN                   & 81.82±0.32          & 38.33±5.02         & 27.58±4.65         & 14.56± 5.44        & 4.00±1.17          & 76.62±2.75          & 30.60±3.95         & 21.37±4.63         & 8.32±1.80          & 4.08±2.32          \\
                          & FairVGNN                  & \textbf{82.45±0.17} & 1.44±2.57          & 0.92±1.10          & 13.04±8.49         & 17.68±24.72        & \textbf{82.81±0.69} & 5.31±5.84          & 0.97±1.39          & 8.24±11.25         & 11.20±9.78         \\
                          & NIFTY                     & 81.25±0.09          & 3.46±1.73          & 4.43±0.80          & 0.48±0.44          & 0.72±0.76          & 77.82±1.45          & 6.67±3.95          & 2.96±3.56          & \textbf{0.32±0.52}         & 0.88±0.44          \\
                          & GRAFair                   & 80.95±0.00     & \textbf{0.81±0.47} & \textbf{0.78±0.56} & \textbf{0.27±0.14} & \textbf{0.68±0.55} & 81.05±0.22          & \textbf{0.28±0.62} & \textbf{0.81±0.92} & 0.98±0.83 & \textbf{0.65±0.58} \\ \hline \hline
\multirow{2}{*}{Datasets} & \multirow{2}{*}{Baseline} & \multicolumn{5}{c|}{Cheb}                                                                                 & \multicolumn{5}{c}{GIN}                                                                                 \\
                          &                           & F1-score($\uparrow$) & $ \Delta_{SP}(\downarrow) $ & $  \Delta_{EO}(\downarrow) $  & $  \Delta_{CF}(\downarrow) $  & $\Delta_{RS}$($\downarrow$)               & F1-score($\uparrow$) & $ \Delta_{SP}(\downarrow) $ & $  \Delta_{EO}(\downarrow) $  & $  \Delta_{CF}(\downarrow) $  & $\Delta_{RS}$($\downarrow$)  \\ \hline
\multirow{5}{*}{Bail}     & Vanilla                   & 76.00±0.88          & 6.92±1.56          & 11.25±2.12         & 13.62±1.94          & 32.05±1.00          & 65.18±9.97          & 9.69±3.25          & 7.52±1.54          & 13.65±4.63         & 24.61±2.12         \\
                          & FairGNN                   & 77.88±0.55          & 5.51±0.86          & 8.47±0.84          & 12.57±0.92          & 22.17±1.20          & 72.63±1.21          & 8.83±1.12          & 7.22±1.19          & 6.77±2.32          & 14.24±1.21         \\
                          & FairVGNN                  & 79.65±0.91          & 2.82±1.87          & 1.81±1.47          & 15.61±2.20          & 18.94±3.76          & 82.02±1.02          & 6.91±0.25          & 6.63±0.51          & 12.83±10.20        & 5.28±0.43          \\
                          & NIFTY                     & 75.78±0.83          & 6.80±1.22          & 10.96±1.77         & 13.32±1.24          & 32.10±0.74          & 72.34±7.22          & 5.18±1.05          & 3.34±1.47          & 1.72±0.24          & 11.60±1.73         \\
                          & GRAFair                   & \textbf{98.34±1.77} & \textbf{1.79±0.44} & \textbf{0.81±0.93} & \textbf{0.39±0.55}  & \textbf{9.98±1.43}  & \textbf{85.78±2.28} & \textbf{1.08±0.30}  & \textbf{1.03±0.79} & \textbf{1.35±0.36} & \textbf{4.84±1.10}  \\ \hline
\multirow{5}{*}{Credit}   & Vanilla                   & 82.13±0.68          & 11.84±3.87         & 9.75±4.01          & 7.02±9.46           & 22.41±4.00          & 80.85±1.02          & 14.44±3.74         & 13.84±3.52         & 24.87±12.22        & 31.52±6.31         \\
                          & FairGNN                   & 81.04±0.24          & 14.91±3.77         & 13.05±4.16         & 11.67±12.25         & 21.30±4.31          & 76.17±1.41          & 13.61±5.90         & 13.53±6.01         & 34.36±20.40        & 27.02±5.18         \\
                          & FairVGNN                  & 82.87±2.56          & 7.36±1.97          & 5.58±1.37          & 5.29±1.66           & 5.88±4.53           & 87.22±0.27          & 0.82±0.55          & 0.66±0.37          & 3.43±3.48          & \textbf{0.78±0.62} \\
                          & NIFTY                     & 82.61±0.88          & 13.69±9.79         & 12.02±9.84         & 11.38±15.49         & 22.94±4.19          & 81.98±2.08          & 8.85±4.82          & 7.75±3.78          & 2.58±3.87          & 7.39±1.78          \\
                          & GRAFair                   & \textbf{85.16±6.30} & \textbf{1.06±1.10} & \textbf{1.77±2.56} & \textbf{2.30±3.24}  & \textbf{1.84±2.84}  & \textbf{87.44±0.17} & \textbf{0.81±0.27} & \textbf{0.53±0.36} & \textbf{2.68±0.82} & 2.23±0.54          \\ \hline
\multirow{5}{*}{German}   & Vanilla                   & 72.96±5.94          & 16.17±6.84         & 9.35±6.17          & 5.60±1.67           & 17.04±3.32          & 82.12±0.93          & 13.30±4.88         & 7.46±4.19          & 6.00±1.60          & 5.60±3.52          \\
                          & FairGNN                   & 80.20±1.13          & 14.89±5.79         & 8.26±4.49          & 3.04±1.46           & 3.84±2.09           & 81.35±3.10          & 18.35±6.66         & 13.03±6.96         & 9.60±2.33          & 4.08±2.73          \\
                          & FairVGNN                  & \textbf{82.28±0.82} & 0.90±2.46          & 1.32±1.10          & \textbf{0.80±10.10} & 2.16±2.21           & \textbf{82.42±0.16} & 1.23±1.78          & 1.09±1.35          & 8.64±12.89         & 17.68±18.69        \\
                          & NIFTY                     & 78.59±6.56          & 14.29±5.34         & 8.13±5.34          & 3.84±1.46           & 12.56±2.15          & 80.60±4.47          & 4.53±8.39          & 6.51±9.68          & \textbf{0.16±0.22}   & 0.48±0.66          \\
                          & GRAFair                   & 80.75±0.92          & \textbf{0.76±0.79} & \textbf{0.92±0.84} & 4.80±10.73         & \textbf{1.60±3.58}  & 80.04±0.79           & \textbf{0.88±0.63} & \textbf{0.72±0.58} & 0.56±0.59 & \textbf{0.44±0.58} \\ \hline
                          
\end{tabular}
\end{table*}

\section{Time efficiency of GNN-based methods}
The time efficiency of fairness-aware GNNs based on different encoders is shown in Table \ref{tab:table_encoder_time}.
\begin{table}[!t]
\small
\renewcommand\arraystretch{1.2}  
\caption{Time efficiency(in seconds) of GNN methods based on different encoders. Each value refers to the average time during training of an epoch. (Bold: the best.)\label{tab:table_encoder_time}}
\centering
\small
\setlength\tabcolsep{4.5pt}
\begin{tabular}{c|c|cccc}
\hline
Datasets                & Baseline & GCN             & GIN             & SAGE            & Cheb            \\ \hline
\multirow{4}{*}{Bail}   & FairGNN  & 0.0351          & \textbf{0.0321}          & 0.0403          & 0.0457          \\
                        & FairVGNN & 0.5295          & 0.0822          & 0.1796          & 0.4285          \\
                        & NIFTY    & 0.0748          & 0.0837          & 0.0936          & 0.0272          \\
                        & GRAFair  & \textbf{0.0210} & 0.0616 & \textbf{0.0198} & \textbf{0.0253} \\ \hline
\multirow{4}{*}{Credit} & FairGNN  & 0.0337          & 0.0390          & 0.0443          & 0.0455          \\
                        & FairVGNN & 0.7119          & 0.1498          & 0.5191          & 1.3182          \\
                        & NIFTY    & 0.0931          & 0.1198          & 0.1260          & 0.0346          \\
                        & GRAFair  & \textbf{0.0254} & \textbf{0.0326} & \textbf{0.0258} & \textbf{0.0276} \\ \hline
\multirow{4}{*}{German} & FairGNN  & 0.0170          & 0.0156          & 0.0184          & 0.0354          \\
                        & FairVGNN & 0.0901          & 0.0384          & 0.1749          & 0.2375          \\
                        & NIFTY    & 0.0433          & 0.0430          & 0.0457          & 0.0234          \\
                        & GRAFair  & \textbf{0.0068} & \textbf{0.0101} & \textbf{0.0084} & \textbf{0.0113} \\ \hline
\end{tabular}
\end{table}

\section{The Detail Implementation}

In this section, we give the hyperparameter of different baselines and GRAFair for their different model architectures.

\textbf{Vanilla GNN}. Learning rate \{0.0001, 0.001, 0.01\}, dropout \{0.0, 0.5, 0.8\}, the number of hidden unit 16.

\textbf{Vanilla w/o S}. \mr{Learning rate \{0.0001, 0.001, 0.01\}, dropout \{0.0, 0.5, 0.8\}, the number of hidden unit 16. Masking the attributes of Race, Age, and Gender on the Bial, Credit, and German datasets respectively.}

\textbf{FairGNN}. Learning rate 0.001, drop edge rate 0.001, drop feature rate 0.1, weight decay 1$\text{e}^{-5}$, hidden size 16, epochs 1000, regularization coefficient 0.6.

\textbf{NIFTY}. Learning rate 0.001, droup out 0.5, weight decay $1\text{e}^{-5}$, hidden size 16, epochs 1000, regularization coefficients $\alpha=4$, $\beta=0.01$.

\textbf{FairVGNN}. Learning rates \{0.001, 0.01\}, dropout 0.5, the number of hidden units 16, the prefix cutting threshold \{0.01, 0.1, 1\}, the whole training epochs \{200, 300, 400\}, regularization coefficient $\alpha \in \{0, 0.5, 1\}$.  

\textbf{Graphair}. Learning rates 0.0001, dropout 0.1, weight decay 1$\text{e}^{-5}$, the number of hidden units 16, training epochs 500, the hyperparameters $\alpha,\beta,\gamma,\lambda \in \{0.1, 1, 1, 10\}$. 

\textbf{FairGT}. Learning rates 0.001, dropout 0.3, weight decay 1$\text{e}^{-5}$, the number of hidden units 64, training epochs 500. 



\textbf{GRAFair}. The model architecture for the node classification task is illustrated in Figure \ref{fig_2}. More
details are listed below:

\begin{itemize}

\item Hyper-parameter $\beta$: \{$10^2$, $5\times 10^2$, $10^3$\}.

\item Learning rate: \{0.001, 0.005, 0.01\}.

\item Backbone GNN models: GCN, GraphSAGE, Cheb and GIN.

\item Training epochs: \{100, 200, 300\}.

\item The number of hidden units: $20$.

\item The number of classifier layers: $\{1, 2\}$.
\item The number of encoder layer: $\{1, 2, 3\}$.

\end{itemize}

\section{datasets statistics}


The statistics of the detailed datasets utilized in the experiment are presented in Table \ref{tab:table5}, providing a comprehensive overview of the data characteristics.

Upon closer examination of the datasets, we can elucidate the bias model inherent within the data. Taking the Credit dataset as an illustrative example, as depicted in Table \ref{tab:5}, a noticeable disproportionality emerges: there exists a significantly higher number of positive samples within the younger age group (age $\leq$ 25) compared to the older age group (age $>$ 25). This observed disparity underscores the presence of bias correlated with sensitive attributes.

Graph Neural Networks (GNNs) trained on such datasets risk perpetuating biases associated with these sensitive attributes. Consequently, GNNs predisposed to age bias may exhibit a tendency to favor positive predictions for younger individuals, notwithstanding the equivalence of other features. Analogous trends are discernible in the Bail and German datasets as well.

\begin{table}[!h]
\small
\setlength{\abovecaptionskip}{0cm}  
\setlength{\belowcaptionskip}{-0.2cm} 
\renewcommand\arraystretch{1.2}  
\caption{Statistics on sensitive attributes of Credit dataset}
\label{tab:5}
\centering
\small
\setlength\tabcolsep{4.5pt}
\begin{tabular}{cc|ccc}
\hline
\multicolumn{2}{c|}{Sensitive attribute}                     & positive & negative & positive ratio \\ \hline
\multicolumn{1}{c|}{\multirow{2}{*}{Age}} & $\leq$ 25              & 21409    & 5906     & 71.36\%                   \\ \cline{2-5} 
\multicolumn{1}{c|}{}                     & \textgreater{}25 & 1955     & 730      & 6.52\%                    \\ \hline
\end{tabular}
\vspace{-1.5em}
\end{table}

\begin{table}[!h]
\small
\renewcommand\arraystretch{1.2} 
\setlength\tabcolsep{4pt}
\centering
\caption{The Datasets statistics\label{tab:table5}}
\begin{tabular}{c|ccc}
\hline
Dataset                                                        & Bail                                                         & Credit                                                                         & German                                                         \\ \hline
Nodes                                                          & 18,876                                                       & 30,000                                                                         & 1,000                                                          \\ \hline
Features                                                       & 18                                                           & 13                                                                             & 27                                                             \\ \hline
Edges                                                          & 321,308                                                      & 1,436,858                                                                      & 22,242                                                         \\ \hline
Average degree                                                 & 34.04                                                        & 95.79                                                                          & 44.48                                                          \\ \hline
\begin{tabular}[c]{@{}c@{}}Sensitive \\ attribute\end{tabular} & \begin{tabular}[c]{@{}c@{}}Race\\ (White/Black)\end{tabular} & \begin{tabular}[c]{@{}c@{}}Age\\ (\textless{}25/\textgreater{}25)\end{tabular} & \begin{tabular}[c]{@{}c@{}}Gender\\ (Male/Female)\end{tabular} \\ \hline
Node label                                                     & Bail decision                                                & Future default                                                                 & Credit status                                                  \\ \hline
\end{tabular}
\end{table}



\bibliographystyle{IEEEtran}
\bibliography{main}

\newpage

 
\vspace{11pt}


\vfill

\end{document}